\documentclass[11pt,a4paper]{article}
\usepackage[T1]{fontenc}
\usepackage[utf8]{inputenc}
\usepackage{lmodern}
\usepackage[margin=1in]{geometry}
\usepackage{microtype}
\usepackage{needspace}
\usepackage{amsmath,amssymb}
\usepackage{graphicx}
\usepackage{booktabs,array,siunitx,float}
\usepackage{tikz}
\usetikzlibrary{arrows.meta,calc,positioning}
\usepackage{fvextra}
\usepackage[numbers,sort&compress]{natbib}
\usepackage{xurl}
\usepackage[hidelinks]{hyperref}
\usepackage{caption}
\usepackage[section]{placeins}
\newcommand{\sourceinput}[1]{%
  \VerbatimInput[fontsize=\footnotesize,breaklines=true,breakanywhere=true,
    breakautoindent=false,breaksymbolleft={},breaksymbolright={},
    frame=single,framesep=2mm,baselinestretch=1.03]{#1}%
}
\hypersetup{pdfauthor={Sergey Rodionov},pdftitle={Improving Environment-Agnostic
Agents: The Role of Scaffolding},
  pdfsubject={Study draft with Results, Discussion, Conclusions, and run-level
  reporting}}

\DeclareRobustCommand{\AgentBaseline}{Baseline}
\DeclareRobustCommand{\AgentStopReport}{Stop-and-report}
\DeclareRobustCommand{\AgentControlledReset}{Controlled\_RESET}
\DeclareRobustCommand{\AgentTwma}{\texttt{twma\_v1.61}}
\newsavebox{\compactmeasure}
\title{Improving Environment-Agnostic Agents:\\
The Role of Scaffolding}
\author{Sergey Rodionov\\[3pt]\small SingularityNET}
\date{}
\begin{document}
\maketitle
\begin{abstract}
Agents built on stronger language models may increasingly construct useful procedures
for themselves, reducing the advantage of developer-supplied scaffolding. We investigate why this
advantage persists and whether an environment-agnostic agent can be improved without
environment-specific engineering. We evaluate a Codex-based agent in two nearly
identical prompt variants on ARC-AGI-3, AGI Maze, and Terminal-Bench. Each variant is
deployed with unchanged code, prompts, and general-purpose tools across environments,
receiving environment-specific information as input. Changing only an instruction to
encourage reporting difficulties and returning control for automatic continuation
improves mean scores on both ARC-AGI-3 and AGI Maze, while Terminal-Bench shows small,
mixed changes. Yet the general agent remains below an ARC-AGI-3-specific reference,
including with GPT-5.6-sol. The reference prescribes organization and control rather
than supplying new game rules or solutions. To investigate this gap, we transfer reset
decisions to a scripted controller that follows a convention already given to the agent.
This small change recovers part of the specialized agent's advantage. A complementary
AGI Maze experiment shows that additional tutorial experience can reduce performance.
Together, these findings highlight the distinction between having information and tools
available and using them effectively. Supplied procedures still matter for the models
tested here. The performance gains also come with higher token-based cost. The
experiments do not separate better use of information and procedures from additional or
differently allocated inference. We provide the EnvTraverse framework and an evaluated
baseline agent for testing whether further environment-independent changes improve
performance in some environments without degrading it in others.
\end{abstract}

\section{Introduction}
\label{sec:introduction}

An LLM-based agent combines a language model with a harness: the prompts, tools, memory
mechanisms, and control logic that support its interaction with an
environment~\cite{bandel2026general}. For this study, we distinguish three components:
the \emph{backbone LLM}; \emph{environment-independent scaffolding}, such as shell
access, a file system, and generic context management; and \emph{environment-specific
scaffolding}, such as controllers and observation layouts designed for a particular
environment family. These components interact. A capable agent may depend heavily on its
harness without requiring that harness to be environment-specific.

Existing research shows that agent performance depends on how models are prompted and
connected to tools. CodeAct investigates executable code as an action representation,
while SWE-agent examines how agent--computer interface design affects
performance~\cite{wang2024codeact,yang2024sweagent}. AgentBench and AgentGym provide
multi-environment evaluation and learning frameworks, and Exgentic compares
general-purpose agent architectures across heterogeneous benchmarks without
per-environment task-solving
customization~\cite{liu2024agentbench,xi2024agentgym,bandel2026general}. Research on
iterative self-refinement and test-time computation further motivates distinguishing the
model from the procedures and inference used to apply
it~\cite{madaan2023selfrefine,snell2024testtime}. We build on these lines of work by
asking which benefits of scaffolding can be retained through mechanisms that remain
unchanged across environments.

Our earlier ARC-AGI-3 studies motivate this question. The first combined executable
world modeling, prompted simplification, and replay verification in a Codex-based
agent~\cite{rodionov2026executableworldmodels}. The second compared textual and
executable-model variants, finding that gains from model capability and reasoning effort
often exceeded differences among those variants~\cite{rodionov2026components}. These
findings are consistent with the hypothesis that stronger models may increasingly
construct useful task-specific procedures themselves, reducing the advantage of
developer-supplied scaffolding. This possibility is related to Sutton's emphasis on
general methods that exploit increasing computation~\cite{sutton2019bitter}. However,
all variants in our earlier studies retained ARC-AGI-3-specific organization and
control, leaving the contribution of that scaffolding unclear.

We therefore ask two related questions. First, \emph{can an environment-agnostic agent
be improved on some environments without material regressions on others?} Second,
\emph{with the underlying model fixed, why can changes in guidance and control improve
performance without supplying new task facts or general-purpose tools?}

We investigate these questions using one Codex-based agent in two nearly identical
prompt variants, each deployed unchanged across ARC-AGI-3, AGI Maze, and Terminal-Bench.
A short instruction encouraging the agent to stop and report difficulties improves mean
scores on ARC-AGI-3 and AGI Maze, while Terminal-Bench shows small changes with no
consistent direction. The focus is on demonstrating and studying a concrete
cross-environment improvement, rather than optimizing the agent for the highest possible
benchmark scores. ARC-AGI-3-specific control provides further gains, including with
GPT-5.6-sol, showing that supplied procedures still matter for the models tested here. A
complementary AGI Maze experiment examines whether the agent benefits from tutorial
experience. Together, these experiments connect a practical improvement to a broader
problem: an agent can have useful information and tools available without reliably using
them effectively.

\paragraph{Scope and approach.}
We use \emph{environment-agnostic} to describe an unchanged agent implementation, not a
guarantee of success in every environment. For a given variant and model--effort
configuration, its code, prompts, and initial agent-owned tools remain fixed across
environments, with no developer-authored environment-specific control logic. Each
environment supplies an environment description and optional files, then accepts JSON
actions and returns observation files through a common interaction contract. These
environment specifications are held fixed in agent comparisons. The agent may construct
specialized tools or strategies during a run; the developer does not supply a different
scaffold for each environment.

Our environment-agnostic \AgentBaseline{} is a simplified version of the
ARC-AGI-3-specific textual agent from our earlier study. A small scripted controller and
an environment-agnostic client provide access to the task, while Codex performs the
reasoning, computation, and action selection. \AgentStopReport{} changes only one
instruction, included in both the initial and continuation prompts. It encourages Codex
to return control when it encounters difficulty or uncertainty. The controller resumes
the same conversation with a reminder of the goal and a suggestion to reread the
environment description when needed, but supplies no human feedback or task-specific
advice. Both variants have access to the same reminder; the modified instruction changes
when Codex is encouraged to return.

We evaluate GPT-5.5 at high and xhigh reasoning effort and GPT-5.6-sol at xhigh and max.
The environments span visual rule discovery in ARC-AGI-3, stateful labyrinths in AGI
Maze, and practical command-line tasks in
Terminal-Bench~2.1~\cite{arcprize2026arcagi3,potapov2026agimaze,merrill2026terminalbench,terminalbench2026v21}.
AGI Maze is evaluated both as independent labyrinths and as sequences of seven tutorials
followed by a target labyrinth.

\paragraph{Findings and contribution.}
\AgentStopReport{} increases mean ARC-AGI-3 Relative Human Action Efficiency (RHAE) by
1.37--7.29 percentage points across the four configurations. It also has higher mean AGI
Maze RHAE70 in seven of eight comparisons, with differences ranging from $-0.20$ to
$+5.23$ percentage points. RHAE70 measures
completion and action efficiency against a chosen fraction of the benchmark's reference
action budget; higher is better. Terminal-Bench shows small, mixed changes. The
game-score gains come with higher token-based cost. They establish improvements under
the tested execution policies, not superior performance at matched inference cost.

To investigate the remaining specialization gap, we use the textual agent from our
earlier ARC-AGI-3 study as a reference, updated to \AgentTwma{}. Its level-and-attempt
organization, reset management, and textual-model instructions prescribe ways of using
information already available to the general agent, rather than supplying additional
game rules or solutions. We choose it instead of the verification variant to avoid the
additional predefined interfaces and programs supplied by that variant. Because
\AgentTwma{} still differs in several respects, we also perform a targeted comparison:
adding scripted RESET management to \AgentStopReport{} recovers approximately 34--60\%
of the observed gap. Codex is instructed not to call RESET, and the controller issues it
at \texttt{GAME\_OVER}. This tests whether a convention already available in the
environment description becomes more useful when implemented explicitly in the
controller. Section~\ref{sec:diagnostic-variants} describes the two ARC-AGI-3-specific
agents.

The AGI Maze tutorial experiment provides complementary evidence about the value of
prior experience. Tutorials reduce RHAE70 for both GPT-5.5-based variants at both effort
settings. With GPT-5.6-sol they help both variants at xhigh, while their effects are
smaller and mixed at max.

Together, the prompt comparison, ARC-AGI-3 diagnostics, and tutorial experiment
highlight the distinction between having useful information and tools available and
using them effectively. The performance gains may reflect more effective information
use, more reliable control, or additional or differently allocated inference. These
effects may act together; the experiments do not separate them.

We introduce EnvTraverse, an experimental framework for running and evaluating agents
across interactive environments, together with the \AgentBaseline{} agent evaluated in
this study. These are intended to support further research on improving
environment-agnostic agents through better procedures or additional general
capabilities. New agent variants can be compared with \AgentBaseline{} across
environments to assess performance gains, possible regressions, and inference cost.

\paragraph{Code and data availability.}
Source code and documentation are available through the EnvTraverse project page at
\url{https://github.com/astroseger/envtraverse}. The complete results and run artifacts
are available through the same page.

\section{Related Work}
\label{sec:related-work}

\subsection{General agents and cross-environment evaluation}
\label{sec:related-general-agents}

General Game Playing studies agents that receive formal descriptions of previously
unspecified games and play without algorithms engineered for each
game~\cite{genesereth2005ggp}. More recent work evaluates or learns language-model
agents across interactive domains: AgentBench provides eight environments, while
AgentGym combines a multi-environment framework, interaction trajectories, and methods
for agent learning and evolution~\cite{liu2024agentbench,xi2024agentgym}. Our study
examines changes to a Codex-based scaffold without additional model training, with each
agent variant deployed unchanged across environments.

Exgentic is the closest cross-environment evaluation precedent. Bandel et al. compare
five agent architectures and five backbone models across six benchmarks, using protocol
wrappers to integrate the original implementations without per-environment task-solving
customization~\cite{bandel2026general}. In their closed-source comparison, backbone
choice explains more aggregate performance variation than architecture, although
architecture remains consequential within a model. The study also includes a controlled
tool-shortlisting ablation and behavioral analysis identifying premature termination
among its failure patterns. We complement this work by using a small prompt change to
demonstrate and study a concrete improvement across environments. The ARC-AGI-3
diagnostics examine the remaining value of supplied procedures when the relevant game
information and general-purpose tools are already available. The AGI Maze tutorial
comparison tests whether additional relevant experience improves performance.

Standardized interaction is also established infrastructure: Gymnasium exposes an
action--observation API, and OpenEnv provides isolated environments with
network-accessible interfaces~\cite{farama2026gymnasium,openenv2026}. EnvTraverse uses a
common interaction contract to separate agent implementations from environment
specifications, which are held fixed when comparing agents.

\subsection{Coding agents and scaffolding}
\label{sec:related-scaffolding}

ReAct interleaves reasoning with actions and environmental feedback~\cite{yao2023react}.
CodeAct uses executable Python as an action representation, supporting tool composition
and revision in response to execution results~\cite{wang2024codeact}. SWE-agent shows
that the commands and feedback provided for repository navigation, file editing, and
execution can materially affect software-engineering
performance~\cite{yang2024sweagent}. StateM organizes execution through persistent
workflow states, runbooks, and checked transitions~\cite{qin2026statem}. These
approaches motivate distinguishing access to computational tools from the organization
and control of their use.

Programmatic world modeling provides examples of constructing and using task-specific
computational aids. WorldCoder builds Python world models from interaction for planning,
while GIF-MCTS generates and repairs code world models from descriptions and collected
trajectories~\cite{tang2024worldcoder,dainese2024codeworldmodels}. In ARC-AGI-3,
OPINE-World combines interactive exploration, object-centric modeling, predictive checks
against recorded transitions, and planning~\cite{courtis2026opineworld}. Tycho studies
when a coding agent should construct, repair, use, or bypass a programmatic
model~\cite{lehmann2026tycho}.

Our earlier study compared textual and executable-model scaffolds with simplification
and verification~\cite{rodionov2026components}. In the present study, the
environment-agnostic variants leave the choice of task-solving models and auxiliary
tools to Codex. We examine a small change to general guidance across environments and
use the ARC-AGI-3 comparisons to study the remaining benefit of supplied organization
and control.

\subsection{Reflection and learning from experience}
\label{sec:related-reflection}

Self-Refine uses the same language model to generate an output, provide feedback, and
revise it without additional training~\cite{madaan2023selfrefine}. Reflexion generates
verbal reflections from task feedback and retains them in memory for subsequent trials,
also without updating model weights~\cite{shinn2023reflexion}. These methods explicitly
organize how feedback is produced and reused. Benefits from reconsideration are not
automatic: Huang et al.\ find that self-correction without external feedback can reduce
performance on the reasoning tasks and models they examine~\cite{huang2023selfcorrect}.

Experience can also be retained as executable behavior. Voyager accumulates reusable
skills through interaction in Minecraft and retrieves them when solving later
tasks~\cite{wang2023voyager}. This provides an example of organizing experience into
computational tools.

These studies motivate examining how agents revisit information and use prior
experience. Our \AgentStopReport{} instruction encourages Codex to return when it
encounters difficulty, after which the controller resumes the conversation with a goal
reminder, without prescribing a separate critique-and-revision procedure. The AGI Maze
comparison separately tests whether preceding tutorial experience benefits the unchanged
agent.

\subsection{Test-time computation}
\label{sec:related-inference}

Research on test-time computation examines how model performance depends on the amount
and organization of inference. Brown et al.\ study repeated sampling, showing that
generating more candidate solutions increases the fraction of problems for which at
least one correct solution is found; benefiting from this increase requires identifying
a correct candidate~\cite{brown2024monkeys}. Snell et al.\ investigate learned answer
revision and verifier-guided search, finding that the relative effectiveness of
inference strategies depends on problem difficulty~\cite{snell2024testtime}.

Computation can also be extended within a single reasoning process. In s1, Muennighoff
et al.\ fine-tune a reasoning model and use budget forcing to control generation length,
including prolonging a reasoning trace when the model attempts to
finish~\cite{muennighoff2025s1}. This provides an example of how an intervention around
stopping can change inference expenditure and performance.

These studies motivate considering both how much inference an agent performs and how it
is organized. In our study, \AgentStopReport{} changes when Codex returns to an external
controller, which resumes the conversation while the environment remains active. Even
with the model and reasoning-effort setting fixed, this may change how long the agent
continues working and when additional inference is used. We therefore examine inference
expenditure as one possible contributor to the observed performance differences.

\section{Environment-Agnostic Framework}
\label{sec:framework}

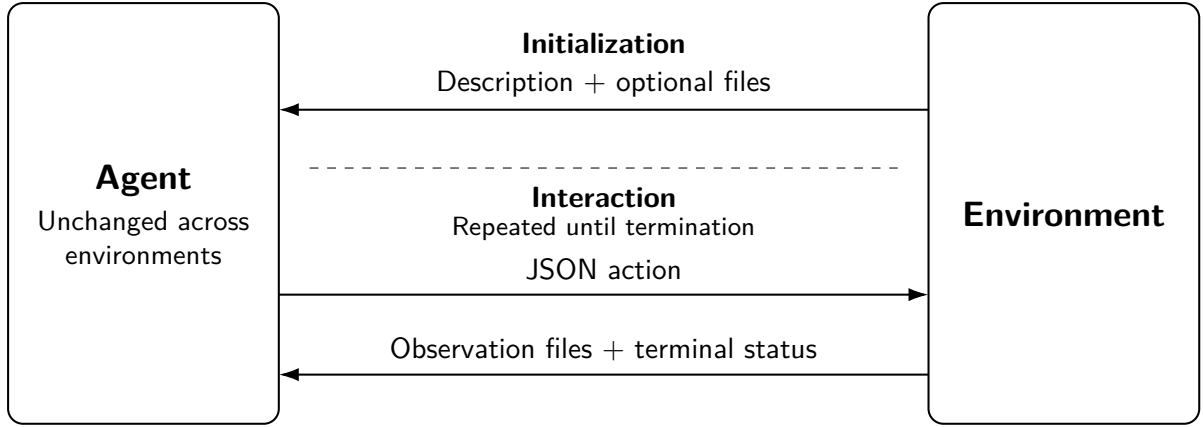
\begin{figure}[t]
\centering
\resizebox{\linewidth}{!}{\begin{tikzpicture}[
  x=1cm,y=1cm,
  font=\sffamily\fontsize{10}{12}\selectfont,
  flow/.style={-{Latex[length=2.4mm,width=1.6mm]},line width=0.7pt},
  phase/.style={font=\sffamily\fontsize{10}{12}\selectfont\bfseries,align=center},
  label/.style={align=center,inner sep=0.5mm}
]
\draw[rounded corners=1.8mm,line width=0.7pt] (0,0) rectangle (3.25,5.05);
\draw[rounded corners=1.8mm,line width=0.7pt] (11.05,0) rectangle (14.30,5.05);
\node[font=\sffamily\fontsize{12}{14}\selectfont\bfseries] at (1.625,2.92) {Agent};
\node[align=center,text width=2.85cm,font=\sffamily\fontsize{9.5}{11.5}\selectfont] at (1.625,2.24) {Unchanged across\\environments};
\node[font=\sffamily\fontsize{12}{14}\selectfont\bfseries] at (12.675,2.525) {Environment};

\node[phase] at (7.15,4.57) {Initialization};
\draw[flow] (11.05,3.77) -- (3.25,3.77);
\node[label] at (7.15,4.08) {Description + optional files};

\draw[dashed,line width=0.35pt] (3.62,3.07) -- (10.68,3.07);
\node[phase] at (7.15,2.72) {Interaction};
\node[font=\sffamily\fontsize{9}{11}\selectfont] at (7.15,2.35) {Repeated until termination};

\draw[flow] (3.25,1.55) -- (11.05,1.55);
\node[label] at (7.15,1.86) {JSON action};
\draw[flow] (11.05,0.59) -- (3.25,0.59);
\node[label] at (7.15,0.90) {Observation files + terminal status};
\end{tikzpicture}}
\caption{Common agent--environment interaction contract. At initialization, the
environment supplies a textual description of the task and interaction interface,
optionally accompanied by files. The agent then submits JSON actions and receives
observation files and terminal status. The agent remains unchanged across environments,
while each environment defines its own actions, observations, and termination
conditions.}
\label{fig:interaction-contract}
\label{fig:architecture} %
\end{figure}

EnvTraverse connects agents and environments through a common interaction contract. Each
environment describes the task and how to interact with it, while the agent determines
how to solve the task. The same agent can be connected to different environments without
modification. New agents and environments can be integrated independently through the
framework's interfaces; agents need not use Codex or an LLM.
Figure~\ref{fig:interaction-contract} summarizes the interaction contract.

\subsection{Interaction contract}
\label{sec:interaction-contract}

\paragraph{Initialization.}
The environment provides an environment description: text explaining the task, the goal
to be achieved, and how the agent can interact with the environment. It may also provide
supplementary files, such as documentation, helper programs, or initial observations,
which are made available in the agent's workspace. We treat these files as part of the
environment description. The environment determines how much information about its rules
and state the description provides.

\paragraph{Actions and observations.}
At each step, the agent submits an action encoded as JSON. The environment defines the
valid action structures and their meaning. In response, it returns an observation
consisting of an arbitrary set of named files. The contract does not prescribe their
formats: observations may contain text, images, or other data. An action may also
reference files in the agent's designated workspace that are accessible to the
server-side infrastructure.

\paragraph{Termination.}
The environment decides when the interaction session ends and indicates this in its
response. Once the session is terminal, no further actions are accepted. Termination
does not necessarily indicate success; its meaning and the criteria for accomplishing
the task are defined by the environment.

After initialization, interaction follows an agent-initiated request--response pattern:
the environment does not send unsolicited observations. This restriction concerns
communication, not whether the environment continues to evolve between requests.

The environment description, action semantics, observation formats, and termination
rules are part of the environment specification. They are held fixed when comparing
agents; changes to them define a new version of the environment specification.

\subsection{Implementation}
\label{sec:runtime}

The implementation runs the agent and the environment server in separate Docker
containers. Actions are JSON objects exchanged over HTTP, and responses carry
observation files and terminal status. Initialization delivers the description text and
an initial observation; supplementary files are placed in the agent's workspace before
interaction begins.

The server-side infrastructure can access the agent's designated workspace, while the
agent has no reciprocal direct access to the task implementation or task sandbox.
Terminal-Bench uses this convention for \texttt{send\_path}: the agent names a source
file or directory and a destination, and the environment copies the requested contents
(see Appendix~\ref{app:terminal-description}).

In the evaluated setup, the agent container accesses the Internet only through an
allowlisted proxy that permits connections to the domains needed by the LLM service.
This restriction does not apply to the environment. An environment may use the Internet
itself or provide Internet access to the agent through its actions. Evaluation-specific
access controls are described in Section~\ref{sec:information-access}. Implementation
documentation is available through the
\href{https://github.com/astroseger/envtraverse}{EnvTraverse project page}.

\section{Environment-Agnostic Agents}
\label{sec:agents}

We study one environment-agnostic agent in two prompt variants: \AgentBaseline{}
(\texttt{agent\_v1}) and \AgentStopReport{} (\texttt{agent\_v1\_01\_wstop}). They use
the same code, controller, and tools; only one short instruction differs. Each variant
is deployed unchanged across environments for a given model--effort configuration. The
agent simplifies the textual agent described in Section~4, ``Architecture and Agent
Variants,'' of our earlier ARC-AGI-3 study~\cite{rodionov2026components}. We retain its
Codex-based architecture while removing ARC-AGI-3-specific control and requirements to
maintain a world model.

\subsection{Agent architecture}
\label{sec:shared-agent}

\begin{figure}[htbp]
  \centering
  \includegraphics[width=0.96\linewidth]{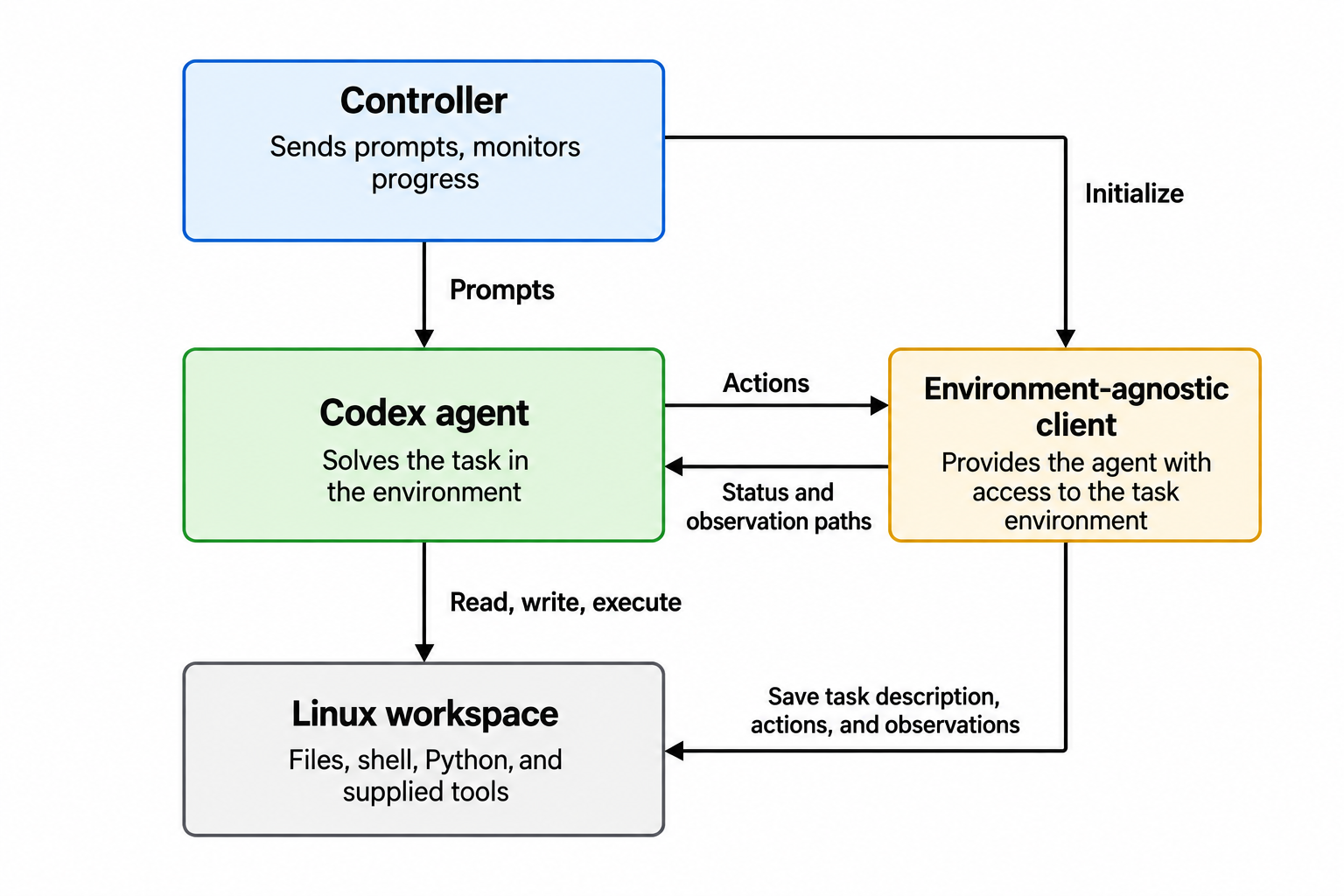}
  \caption{Architecture shared by the two prompt variants. The scripted controller sends
  prompts and checks generic session state. Codex performs the task-solving work and
  uses the provided client to interact with the environment. The client is part of the
  agent and saves the environment description, actions, and observations. All components
  remain unchanged across environments.}
  \label{fig:agent-architecture}
\end{figure}

Figure~\ref{fig:agent-architecture} shows the agent's four components: a controller, a
Codex coding agent, an environment-agnostic client, and a persistent Linux workspace.
Codex performs the task-solving work: it reads observations, reasons about the task,
writes and executes programs, and selects environment actions. It submits those actions
through the provided client, which is part of the agent rather than the environment. The
client provides the same interface across environments and contains no
environment-specific logic.

The main prompt (Appendix~\ref{app:main-prompt}) asks Codex to read the environment
description and accomplish the goal stated there. It explains how to invoke the client
with a JSON action and where to find the resulting observations. The client saves the
description, submitted actions, and returned files, and prints their locations.
Observation records use a generic step-indexed layout rather than an
environment-specific organization.

Codex can use files, the shell, Python, and supplied tools in its workspace. The main
prompt asks it to use the workspace as it finds most effective and suggests helper
indexes, access scripts, or reorganized copies of observations when useful, while
prohibiting changes to the original observation records
(Appendix~\ref{app:main-prompt}). It does not require a world model, planner, or
verifier; Codex decides which additional files and tools to create. A single Codex
invocation can include many model calls, local computations, and environment actions
before returning control; it is not a single LLM call or environment step. The workspace
and Codex conversation persist across ordinary continuations within a run.

\subsection{Controller}
\label{sec:continuation}
\begingroup
\clubpenalty=10000
\widowpenalty=10000

The controller is a small script that sends prompts and checks basic session state; it
contains no task-solving logic. It follows the flow in
Figure~\ref{fig:agent-controller}. First, it initializes the environment through the
client, then sends the main prompt, the variant-specific continuation instruction, and
the initial client output identifying the description and observation files.
Appendix~\ref{app:prompt-composition} gives the exact composition. The instruction
distinguishing the variants is therefore present from the first invocation.

\begin{figure}[htbp]
  \centering
  \includegraphics[width=0.94\linewidth]{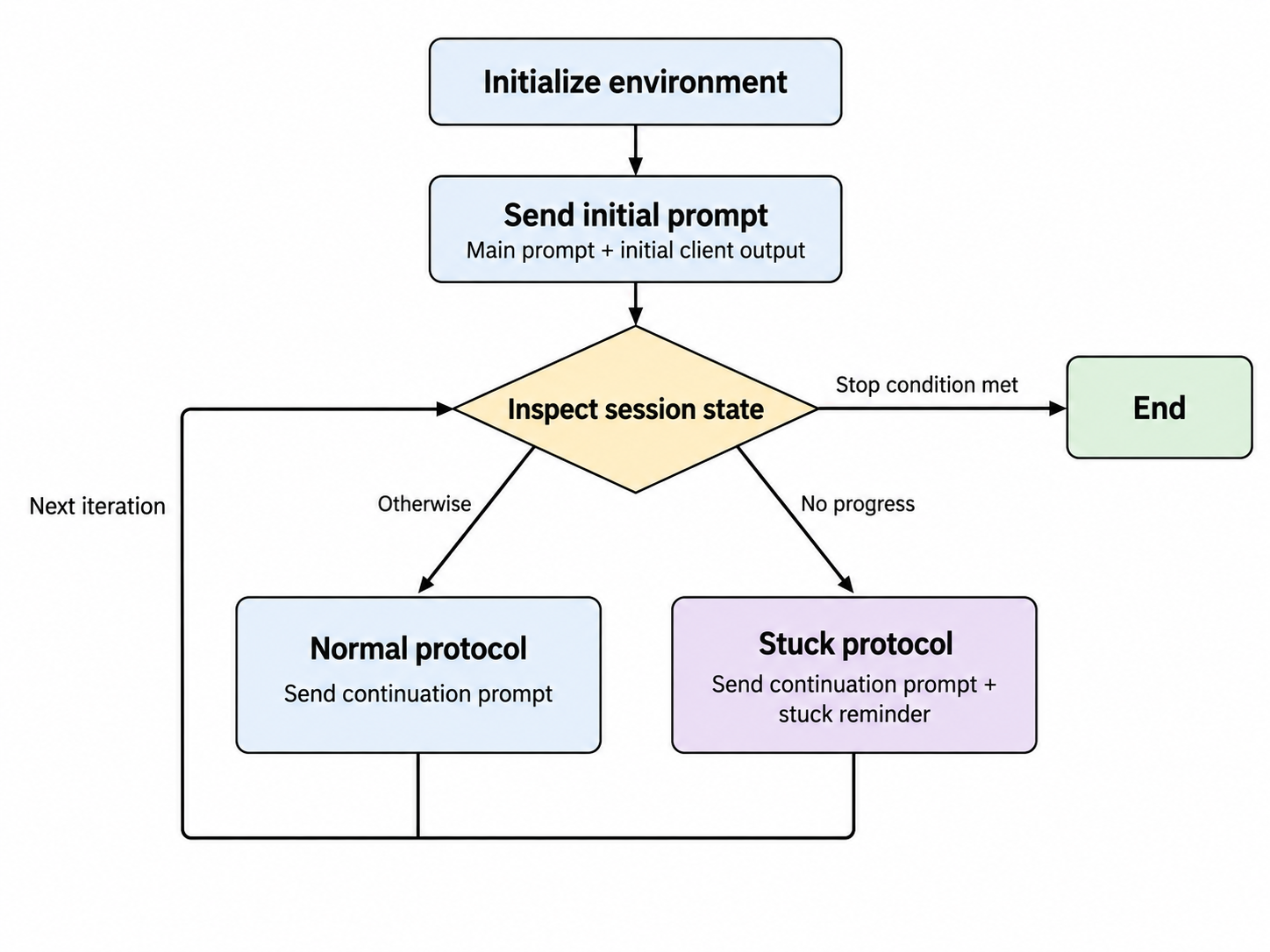}
  \caption{Scripted controller used by both prompt variants. The initial prompt includes
  the selected continuation instruction. State is inspected after Codex returns; ``No
  progress'' means no new action since the preceding continuation. ``End'' denotes
  controller exit after environment termination or a return with no new action following
  the stuck reminder. The runtime-recovery supervisor is omitted; its handling of
  controller exits is described in the text.}
  \label{fig:agent-controller}
\end{figure}

When Codex returns, the controller reads the terminal flag and latest environment-step
index from saved client state. It does not interpret observations or assess Codex's
reasoning. If the environment is terminal, the controller exits; otherwise, it selects
one of two protocols.

\paragraph{Normal protocol.}
The controller resumes the same Codex conversation with a reminder to keep working
toward the goal in the environment description (Appendix~\ref{app:continuation-prompt}),
followed by the variant-specific instruction. The first nonterminal return receives this
continuation even if the initial invocation took no action.

\paragraph{Stuck protocol.}
If a continuation produces no new environment action, the controller adds the no-action
reminder (Appendix~\ref{app:stuck-reminder}). If Codex again returns without taking an
action, the controller exits. If the environment remains nonterminal, this exit triggers
recovery, subject to the recovery limit described below. This is a fallback for repeated
non-action.

\paragraph{Recovery.}
In our previous experiments, Codex occasionally exited with a runtime or service error,
or appeared to freeze while its process remained alive but stopped producing observable
activity~\cite[Section~4.7, ``Runtime recovery'']{rodionov2026components}. Following
that work, we use a recovery supervisor to continue execution after such failures.

The supervisor monitors the controller process and its activity log. When the process
exits or the log remains inactive beyond the configured threshold, it restarts the
controller if saved session state exists and the environment is still active. Recovery
preserves the environment, workspace, and recorded observations. It normally resumes the
saved Codex conversation, unlike the earlier implementation, which started a new
conversation.\footnote{In our implementation, recovery starts a new Codex conversation
if no environment action has yet been taken in the run; otherwise, it resumes the saved
conversation.} Implementation details are provided in the source code and documentation
available through the \href{https://github.com/astroseger/envtraverse}{EnvTraverse
project page}.

As a consequence of the same restart rule, the supervisor also restarts the controller
after an exit following the stuck reminder. It permits up to ten recovery attempts in
total per environment run, after the initial launch. Ordinary Codex returns and normal
continuations do not count against this allowance. The repeated opportunities after the
stuck protocol are therefore a consequence of the recovery mechanism, whose main purpose
is to handle technical Codex failures.

\endgroup

\subsection{Prompt difference and expected lifecycle}
\label{sec:return-instruction}

The sole difference between the variants is the text of
\texttt{continuation\_string.txt}, included at initialization and in every continuation.
For \AgentBaseline{}, this instruction is:
\begin{quote}
Continue working until you have completed your task.
\end{quote}
For \AgentStopReport{}, it is:
\begin{quote}
Continue working on the task as long as you can proceed confidently. If you encounter a
problem, ambiguity, unexpected issue, or anything that makes you unsure how to proceed,
stop and report it back to me before taking further action.
\end{quote}
The complete source strings are reproduced in Appendices~\ref{app:baseline-instruction}
and~\ref{app:stop-instruction}.

\paragraph{\AgentBaseline{} lifecycle.}
Consider a successful run in which Codex follows the \AgentBaseline{} instruction. The
controller sends the initial message, and Codex carries out the task until the
environment becomes terminal. Codex then returns and the controller exits. Neither the
normal nor the stuck protocol is entered in this successful path: all task-solving work
occurs within the initial Codex invocation. If Codex returns before termination despite
this instruction, the controller prompts it to continue.

\paragraph{\AgentStopReport{} lifecycle.}
The initial message instead asks Codex to return when it encounters a difficulty or
uncertainty. If this happens before termination, the controller proceeds directly to the
normal protocol and tells Codex to continue using the continuation text and the same
\AgentStopReport{} instruction. The conversation and workspace are retained. Here,
``report it back to me'' refers to the automated controller, not a human: the controller
does not answer the report or supply a task-specific suggestion. Further returns follow
the normal or stuck protocol described above. This variant can also finish the task
without returning early.

The intended difference is therefore when the same task-solving process returns control,
not a change in its tools or environment-specific capabilities. The instructions request
these behaviors rather than enforcing them; their effects on performance and resource
use must be evaluated from the runs.

\section{Implemented Environments}
\label{sec:environments}

We build environments from three existing benchmarks: ARC-AGI-3, a collection of
interactive visual puzzles~\cite{arcprize2026arcagi3}; AGI Maze, a framework of stateful
grid labyrinths~\cite{potapov2026agimaze}; and Terminal-Bench~2.1, a suite of practical
command-line
tasks~\cite{merrill2026terminalbench,terminalbench2026v21}.\footnote{Project websites:
ARC-AGI-3, \url{https://arcprize.org/arc-agi/3}; AGI Maze, \url{https://agimaze.org/};
Terminal-Bench~2.1, \url{https://www.tbench.ai/news/terminal-bench-2-1}.} Terminal-Bench
provides a substantially different setting from the two game families: the agent works
on files and programs in a task sandbox, and processes can continue executing between
actions.

The broad contract in Section~\ref{sec:framework} allows these different tasks to be
exposed through the same interface. We distinguish the underlying \emph{benchmark} from
the \emph{environment} defined for our experiments. An environment includes its
environment description, comprising text and optional files; the dynamics that accept
JSON actions, reject invalid requests, and return observation files; and its termination
conditions. These choices define the environment, rather than merely documenting its
implementation. Changing the ARC-AGI-3 description, omitting its PNG observations, or
adding a direct \texttt{exec} action to Terminal-Bench would therefore produce a
different environment specification, even with the same underlying task. Comparisons
between agents keep the specification fixed.

The following subsections describe these three elements for each setting.
Appendix~\ref{app:environment-inputs} reproduces the information supplied to the agent;
we refer to those descriptions for the full instructions and action syntax rather than
repeat them here.

\subsection{ARC-AGI-3}
\label{sec:arc-environment}

One environment contains a single ARC-AGI-3 game, including its successive levels and
attempts.

\paragraph{Environment description.}
The same general description is supplied for every game. It explains the objective,
action conventions, observation files, and reset behavior, but provides no game-specific
mechanics or solution. It also encourages visual inspection of the observations.
Appendix~\ref{app:arc-description} gives the complete description and identifies the
supplied rendering helper.

\paragraph{Dynamics and observations.}
The agent selects among the available game actions and receives the resulting visual
frames and metadata. The full file inventory is given under ``Provided Artifacts'' in
Appendix~\ref{app:arc-description}. The important representational choice is to provide
each frame both as exact ASCII color-index data and as a corresponding PNG image,
including intermediate frames when present. This lets the agent process the frame
numerically or inspect it visually, consistent with the visual nature of the task. The
initialization files also include \texttt{ascii\_to\_png.py}, which the agent can use to
render reconstructed or transformed frames. Supplying only ASCII, with or without a
rendering helper, would be another environment version; we do not compare those
representations here.

\paragraph{Termination.}
The environment ends at \texttt{WIN} or when the current-level action counter exceeds
1,500. The counter includes \texttt{RESET} actions, spans all attempts at the same
level, and is cleared when a new level is reached. We use the 1,500-action threshold
discussed in Section~5.4, ``Interaction rules and stopping conditions,'' of our earlier
study~\cite{rodionov2026components}. That budget is an evaluation choice, not a native
ARC-AGI-3 limit, and may restrict games requiring longer exploration. A larger budget
could be used in an evaluation aimed at maximizing performance, but would change the
environment specification.

\subsection{AGI Maze: single labyrinth}
\label{sec:maze-environments}
\label{sec:maze-single-environment}

One environment contains a single labyrinth. AGI Maze already supplies rules, task
information, action identifiers, and textual feedback, so only a thin interface layer is
needed.

\paragraph{Environment description.}
The environment description combines AGI Maze's own rule description with the
labyrinth's public task information, available actions, and public map. All of this
information is provided by AGI Maze. Appendix~\ref{app:maze-single} gives the full
template and a rendered example.

For tutorial labyrinths, the public map reveals all internal walls and features,
including the positions of keys, treasure, rivers, and pits. We use tutorial labyrinths
only in the curriculum-learning experiments with the labyrinth-set environment. For
non-tutorial labyrinths, the public map shows only the external boundary and the
player's initial position, visualizing information already provided in the text.

\paragraph{Dynamics and observations.}
Actions encode the labyrinth-defined choices as JSON objects. After each action, the
environment returns \texttt{state.txt}, containing textual feedback and inventory
information.

\paragraph{Termination.}
The environment ends when the labyrinth is solved or the agent uses five times the
\texttt{max\_steps} reference budget provided by AGI Maze.

\subsection{AGI Maze: labyrinth set}
\label{sec:maze-set-environment}

One environment contains an ordered sequence of labyrinths. The agent remains in the
same run throughout the sequence, allowing it to use experience from earlier labyrinths
when solving later ones.

\paragraph{Environment description.}
The environment description asks the agent to solve all labyrinths efficiently and
states that the first few are tutorials with an open map showing the player's position;
later maps are hidden. Appendix~\ref{app:maze-set-description} reproduces it verbatim.

\paragraph{Dynamics and observations.}
The agent starts each labyrinth with \texttt{\{"id":"start"\}}. The response includes
\texttt{labyrinth\_description.md}, containing the environment description used in the
single-labyrinth setting, and \texttt{state.txt}, containing the same state information
as in the single-labyrinth environment with the labyrinth index added. Tutorial
labyrinths additionally return the current map and player position as \texttt{map.txt};
non-tutorial labyrinths omit this file. This map is also supplied by AGI Maze.

Each labyrinth receives the same fivefold action budget as in the single-labyrinth
setting. Solving it or exhausting that budget produces \texttt{message.md}, reporting
the outcome. When another labyrinth remains, the environment then waits for the next
\texttt{start} action. This separates feedback about the previous labyrinth from
initialization of the next. In particular, the failure message asks the agent to
consider what went wrong before continuing. These messages are fixed environment inputs,
and Appendix~\ref{app:maze-set-transitions} presents their exact text by the situation
in which each is sent.

\paragraph{Termination.}
The set ends only when the final labyrinth is solved or reaches its action limit.
Failure on an earlier labyrinth does not terminate the set: the agent proceeds to the
next one. A terminal set therefore means that the sequence is complete, not that every
labyrinth was solved.

\subsection{Terminal-Bench 2.1}
\label{sec:terminal-environment}

One environment contains a single Terminal-Bench task. Unlike the turn-based game
settings, it can evolve while the agent deliberates. The task specifies a wall-clock
time limit for solving.

\paragraph{Environment description.}
The environment description is generated from the template in
Appendix~\ref{app:terminal-description}, inserting the public task instruction as
\texttt{\{instruction\}} and the solving budget in seconds as \texttt{\{budget\}}. The
template gives the complete action interface and timing conventions. The environment
description also explains that the task sandbox has a separate filesystem from the
agent's workspace and that task files are accessed through environment actions.

\paragraph{Dynamics and observations.}
The interface exposes six action categories. \texttt{terminal\_send} submits literal
text and named keys to an interactive terminal; \texttt{terminal\_read} captures its
screen or a bounded portion of its history; and \texttt{terminal\_new} opens another
independent shell on the same task filesystem. \texttt{get\_file} retrieves a regular
file, while \texttt{send\_path} transfers a file or directory from the agent's workspace
to the task sandbox. Finally, \texttt{finish} submits the current task state and ends
solving. Responses contain terminal snapshots, metadata, transfer acknowledgments, or
retrieved file contents, depending on the action.
Appendix~\ref{app:terminal-observations} lists the returned files.

Terminal input does not wait for command completion. Processes continue between actions,
and an empty or unchanged screen does not imply that a command has finished. A read may
include a short fixed wait before capturing output, but it does not wait for a process
to terminate. The agent must decide when to inspect output, send further input,
interrupt a process, or work elsewhere. Thus, command execution is nonblocking with
respect to completion, while observations are still delivered only in response to
requests.

We do not expose a dedicated execute-and-wait action. This is an interface choice, not a
restriction of Terminal-Bench: its evaluation infrastructure supports direct command
execution by custom agents~\cite{harbor2026customagents}. An \texttt{exec} action with a
short waiting limit would also fit our contract, but would define another environment
version. Whether it helps is outside this study. Here Terminal-Bench tests the unchanged
agent in a setting markedly different from the games, rather than serving as an
interface-optimization exercise.

\paragraph{Termination.}
Calling \texttt{finish} or reaching the task's external timeout ends the solving phase
and prevents further actions. In both cases, the verifier is run and its reward
determines the task score; successful verification can therefore receive reward 1 after
an agent timeout. If the verifier itself times out, the score is zero. Termination and
task success remain distinct: calling \texttt{finish} is not itself evidence that the
task was solved.

\section{ARC-AGI-3-Specific Diagnostic Variants}
\label{sec:diagnostic-variants}

These experiments investigate why an ARC-AGI-3-specific scaffold can outperform the
general agent when the relevant environment information is already available to both. We
first compare against \AgentTwma{}, the updated textual agent from our earlier study,
whose specialization primarily concerns how information is organized and how the
interaction is controlled. Because that comparison involves several differences, we also
evaluate a separate variant, \AgentControlledReset{}, which adds an ARC-AGI-3-specific
reset policy to \AgentStopReport{}. Both diagnostic systems are evaluated only on
ARC-AGI-3, at the same model--effort configurations as the environment-agnostic
variants.

\subsection{Specialized reference: \AgentTwma{}}
\label{sec:twma-reference}

\AgentTwma{} retains the textual agent, \texttt{twma\_v1.6}, described in Section~4,
``Architecture and Agent Variants,'' of our earlier study~\cite{rodionov2026components}.
Its task-solving prompts and controller logic are unchanged. We update Codex to version
0.154.0 and use the same runtime memory safeguards as the environment-agnostic agents in
this study.\footnote{The earlier \texttt{ulimit -v 6000000} is replaced by Docker
settings \texttt{--memory 10g}, \texttt{--memory-swap 10g}, and \texttt{--ulimit
data=2147483648:2147483648}.} The reference is rerun in the present campaign rather than
represented by historical scores.

Two principal structural adaptations are its organization of observations by level and
attempt and its controller-managed RESET policy. The general agent already receives a
description of levels, attempts, and the reset convention
(Appendix~\ref{app:arc-description}). Its main prompt explicitly permits creating
indexes, scripts, and alternative organizations of observations in its workspace
(Appendix~\ref{app:main-prompt}). These adaptations therefore supply organization and
control around available information rather than additional facts about the game.

Other differences concern how Codex is instructed to work. In particular, \AgentTwma{}
requires a persistent textual world model and uses a more structured prompt sequence.
The general agent can also create and maintain such a description, but is left to decide
whether and how to do so. Our motivation is to examine the value of prescribing useful
procedures instead of leaving the agent to construct them. The claim that \AgentTwma{}
supplies no additional game rules or solutions is a qualitative comparison of the
supplied game information, not a claim of identical prompts or control logic.

We choose the textual rather than the verification variant from the earlier study
because the latter additionally supplies predefined model interfaces, templates, and
verification programs~\cite{rodionov2026components}. These are ready-made procedures,
not hidden game solutions, but they introduce further differences in the tools initially
supplied to the agent. \AgentTwma{} makes the distinction between available task
information and prescribed organization easier to examine. Nevertheless, its differences
from the general agent remain bundled, so this comparison does not identify the
contribution of any single component.

\subsection{Targeted specialization: \AgentControlledReset{}}
\label{sec:controlled-reset}

To obtain a more controlled comparison, we start from \AgentStopReport{} and add only
the ARC-AGI-3-specific reset-management intervention. We call this variant
\AgentControlledReset{} (\texttt{test\_arc3\_agent\_v1\_02\_wstop\_wreset}). Its main
prompt adds:
\begin{quote}
If the game is in the \texttt{GAME\_OVER} state, you should report it, and I will give
you another attempt on the same level. You are forbidden to call RESET by yourself.
\end{quote}
The prohibition is a prompt instruction; the client's action interface remains
unchanged.

When Codex returns in \texttt{GAME\_OVER} and the evaluation stopping conditions have
not been met, the controller issues RESET through the client. It then resumes the same
Codex conversation with the normal continuation prompt, a short reset notice, and the
resulting client output. The reset observation is therefore available before Codex
continues. This branch adds RESET before ordinary continuation; unlike the earlier
textual agent's reset protocol, it adds no separate failure-analysis stage.

The \AgentStopReport{} instruction, tools, and step-indexed observation layout are
retained. In particular, the intervention does not introduce \AgentTwma{}'s
level-and-attempt organization or its textual-model requirement. It tests the effect of
transferring reset management from the coding agent to a script, using a convention
already stated in the environment description. It is environment-specific because the
controller interprets \texttt{GAME\_OVER} and applies ARC-AGI-3 reset semantics.

Restricting reset decisions is not claimed to be optimal. If an agent recognizes that an
attempt has become unrecoverable before \texttt{GAME\_OVER}, an early RESET could save
actions otherwise spent reaching the failure state. This is a possible benefit of
agent-managed resets, not an effect evaluated here; the current description advises
avoiding RESET outside \texttt{GAME\_OVER}. The experiment asks whether explicitly
implementing the stated convention improves the present agent. Performance and
resource-use differences are reported in Section~\ref{sec:results-arc3}.

\section{Evaluation Setup and Protocol}
\label{sec:experimental-protocol}

This section specifies the task collections, scoring rules, execution conditions, and
evaluation-block exclusion and replacement policies.

\subsection{Evaluation design}
\label{sec:evaluation-design}

We compare \AgentBaseline{} and \AgentStopReport{} using GPT-5.5 at \texttt{high} and
\texttt{xhigh} reasoning effort and GPT-5.6-sol at \texttt{xhigh} and \texttt{max}.
\AgentControlledReset{} and \AgentTwma{} are additionally evaluated on ARC-AGI-3 at the
same four configurations. Within each benchmark, the environment specification and
configured execution conditions are held fixed across the compared agents.

The \AgentStopReport{} modification was developed through ARC-AGI-3 experimentation,
while preserving the environment-agnostic implementation. AGI Maze performance was not
used to select this instruction; its benefit on AGI Maze was identified in the analysis
reported here.

Following our earlier study~\cite{rodionov2026components}, an \emph{evaluation block} is
one complete task collection evaluated with a specified agent, model, and reasoning
effort. Three evaluation blocks were completed per experimental condition. After the
exclusions in Section~\ref{sec:apps-search-exclusions}, five conditions retain two blocks and all
others retain three. Individual tasks within a block use separate environment runs,
with fresh workspaces and no conversation or task
artifacts carried over between them. Persistence within an AGI Maze curriculum set is
intentional. Repetition indices identify runs, not matched random seeds across agents.

\paragraph{ARC-AGI-3.}
A block contains the 25 public games, with one independent playthrough per game.

\paragraph{AGI Maze.}
A block contains all labyrinths with identifiers \texttt{0005} or \texttt{0009} in AGI
Maze's \texttt{CLASSIC} and \texttt{EXTENDED} collections~\cite{potapov2026agimaze},
giving 106 targets. Each identifier contributes 53 labyrinths. The evaluation budget
allowed two such batches with three repetitions each; we chose the two identifiers
arbitrarily. In the single-labyrinth setting, each target is evaluated alone. In the
curriculum setting, each independent run contains the seven tutorials
\path{TUTORIAL/STAGE-01/0000.json} through \path{TUTORIAL/STAGE-07/0000.json}, in that
order, followed by one target. Only the target contributes to the score and completion
count; resource use includes the tutorials. Each block is executed as separate 53-target
\texttt{0005} and \texttt{0009} batches and paired by repetition index for analysis.
Individual results for both batches are reported in Appendix~\ref{app:maze-individual}.

\paragraph{Terminal-Bench.}
A block contains all 89 Terminal-Bench~2.1 tasks, with an independent task sandbox for
each run.

\subsection{Performance measures and resource use}
\label{sec:evaluation-measures}

For each setting, we report the mean block score and cost over the retained repetitions.
Appendix~\ref{app:individual-results} provides the individual runs. Contrasts are
descriptive differences between means; their direction alone does not establish
statistical significance or non-inferiority.

\paragraph{ARC-AGI-3.}
The block score is the mean benchmark-reported Relative Human Action Efficiency (RHAE)
across all 25 games. RHAE combines level completion with action efficiency relative to
the benchmark's reference action
counts~\cite{arcprize2026arcagi3,rodionov2026components}.

\paragraph{AGI Maze.}
We use \emph{RHAE70}, an RHAE-inspired score with a reference action count equal to 70\%
of AGI Maze's \texttt{max\_steps}. For target $i$, let $m_i$ be its benchmark-provided
\texttt{max\_steps} value and $a_i$ the number of actions used. Its score is
\begin{equation}
 s_i =
 \begin{cases}
  0, & \text{if the target is not solved},\\[2pt]
  100\min\!\left\{1,\left(\dfrac{0.7m_i}{a_i}\right)^2\right\}, & \text{otherwise}.
 \end{cases}
 \label{eq:rhae70}
\end{equation}
The block score is the mean of the per-target scores over all 106 targets. Higher is
better. According to the AGI Maze team, the benchmark-provided \texttt{max\_steps}
budgets include some slack. We chose the factor 0.7 based on the team's internal
playthrough results, summarized in Appendix~\ref{app:rhae70-reference}. This is a
pragmatic reference threshold, not a calibrated human baseline. Tables also report
\emph{x1 solved}, the number of targets solved within $m_i$ actions, as descriptive
completion information. The individual-run appendix additionally retains the mean step
fraction, assigning 5 to unsolved targets.

\paragraph{Terminal-Bench.}
The block score is the mean task-verifier reward over all 89 tasks, shown as a
percentage. Verifier timeouts receive zero reward.
Appendices~\ref{app:tb-agent-timeouts} and~\ref{app:security-errors} report agent
timeouts and cybersecurity-related errors.

\paragraph{Cost tokens.}
We summarize inference expenditure in millions of normalized cost tokens (M), using the
same convention as our earlier study~\cite{rodionov2026components}:
\begin{equation}
 C = 10^{-6}\!\left(\frac{N_{\mathrm{cached}}}{60}
 + \frac{N_{\mathrm{input}}-N_{\mathrm{cached}}}{6}
 + N_{\mathrm{output}}\right),
 \label{eq:normalized-cost}
\end{equation}
where $N_{\mathrm{input}}$ includes cached input. The pricing-inspired weights are a
conventional proxy for inference expenditure, not calibrated ratios of computational
work. They are applied consistently and do not incorporate subscription charges or
service-tier billing. Reported costs cover retained evaluation blocks, including
tutorials and within-run recovery.

\subsection{Preventing unintended information access}
\label{sec:information-access}

Coding agents can inspect files and execute programs, making unintended information
channels relevant to evaluation. Our first article documents such channels in earlier
development harnesses, including access to public game identifiers and apparent use of a
second local client~\cite{rodionov2026executableworldmodels}. The present setup
separates the agent from the environment implementation using Docker containers,
restricts external connections to an allowlisted model-service proxy, and disables Codex
native web search. Codex runs with its own sandbox and approval prompts disabled; separation
from the task implementation is provided by the surrounding container. Before an
independent run, account-local Codex state is cleared apart from authentication and the
minimal configuration specifying search and service-tier settings. Concurrent workers
used separate local Codex state directories, including workers authenticated through the
same subscription.

The launcher replaces the private task identifier with a random public alias accepted by
the environment server. For ARC-AGI-3, this removes the explicit public game name from
the initialization interface as a cue for lookup or retrieval of memorized information.
The server permits only one initialization of its configured environment.

These measures are engineering controls, not a formal security guarantee. A capable
agent may seek other channels or recognize a public task from its observations.
Agent-side network restrictions are also distinct from capabilities intentionally
exposed by a task sandbox. An Apps-mediated search route remained available despite
these controls; all blocks with recorded attempts through that route are excluded, as
described in Section~\ref{sec:apps-search-exclusions}.

\subsection{Apps-mediated web search and block exclusions}
\label{sec:apps-search-exclusions}

In the tested Codex \texttt{0.154.0} configuration,
\texttt{web\_search="disabled"} disabled native web search, but web search remained
available through the \texttt{search\_service.web\_run} tool on the
\texttt{codex\_apps} server. The agent could therefore retrieve web content despite
our network restrictions. We subsequently disabled Codex Apps entirely by setting
\texttt{features.apps=false}.

An audit of the preserved execution logs identified attempted calls to this tool in
15 of the 8,424 task executions (0.18\%) in the completed evaluation blocks.
These attempts occurred in five blocks. Every such block was excluded in full from
the aggregate analysis, whether or not the call
succeeded and irrespective of task outcome. A full rerun was not feasible within our
computational budget. These post-hoc exclusions are separate from the technical-failure
replacement policy in
Section~\ref{sec:failure-policy}; the excluded blocks were not replaced.
Appendix~\ref{app:web-search-exclusions} gives the detection rule, affected tasks,
and excluded blocks.

\subsection{Computational environment and runtime}
\label{sec:computational-environment}

The campaign used two identically configured Linux virtual machines, each with 32 vCPUs
reported by the guest as AMD EPYC 9454 and 64~GiB RAM, matching the hardware
configuration of our earlier study~\cite{rodionov2026components}. Docker was version
\texttt{29.4.3}, and all agents used Codex \texttt{0.154.0}. The reported experiments
ran from 21 September to 10 October 2026 using eight subscription-backed Codex accounts;
this interval excludes exploratory experiments not reported here.

ARC-AGI-3 and AGI Maze used the default service tier, with at most 16 concurrent agents
per server across the two benchmarks combined. They had no additional hard wall-clock or
token cap; the task limits and the limit on recovery attempts still applied.
Terminal-Bench used at most four concurrent agents per server, with no other
experimental workloads on that server, and \texttt{service\_tier = "fast"}. These
hardware, concurrency, and service-tier settings were fixed within each benchmark
comparison. They standardize the configured conditions, not instantaneous remote-service
latency. Because Terminal-Bench is wall-clock limited, we report its service tier
explicitly; we did not separately measure its effect on scores.

\subsection{All-or-nothing policy for irrecoverable failures}
\label{sec:failure-policy}

Technical interruptions can affect trajectories unevenly. On the same task, an agent
that struggles may run longer than one that quickly finds a solution, giving it more
opportunity to encounter a service interruption or host failure. Replacing only
interrupted task runs could therefore preferentially replace difficult trajectories and
bias the reported performance upward. Following our earlier
study~\cite[Section~5.5]{rodionov2026components}, we use an all-or-nothing policy at the
evaluation-unit level.

If an irrecoverable technical failure prevents an evaluation unit from being completed,
the entire unit is discarded and rerun from scratch, including any tasks that had
already finished. For ARC-AGI-3 and Terminal-Bench, the unit is a complete evaluation
block. For AGI Maze, the policy applies separately to the 53-target \texttt{0005} and
\texttt{0009} execution batches. No task result from a discarded unit is reused in its
replacement.

During this campaign, we replaced one accidentally interrupted AGI Maze batch, three
Terminal-Bench blocks in which a server-container startup timeout prevented some tasks
from starting, and one \AgentTwma{} block affected by a model-service capacity error
before the first action of a game. For \AgentTwma{}, whose inherited runner does not
recover failures before the first action, the capacity error was classified as an
operational startup failure.

\clearpage

\section{Results}
\label{sec:results}
\suppressfloats[t]
\clubpenalty=10000
\widowpenalty=10000

We examine the cross-environment effects of the prompt modification, the remaining
advantage of ARC-AGI-3-specific scaffolding, and the effect of tutorial experience in
AGI Maze. Evaluation blocks, scores, and execution policies are defined in
Section~\ref{sec:experimental-protocol}. Tables report means over the retained
repetitions: two for the five conditions affected by block exclusions and three for
all others. Individual runs are given in Appendix~\ref{app:individual-results}.

\subsection{ARC-AGI-3: general improvements and the specialization gap}
\label{sec:results-arc3}

\begin{figure}[tbp]
\centering
\includegraphics[width=\linewidth]{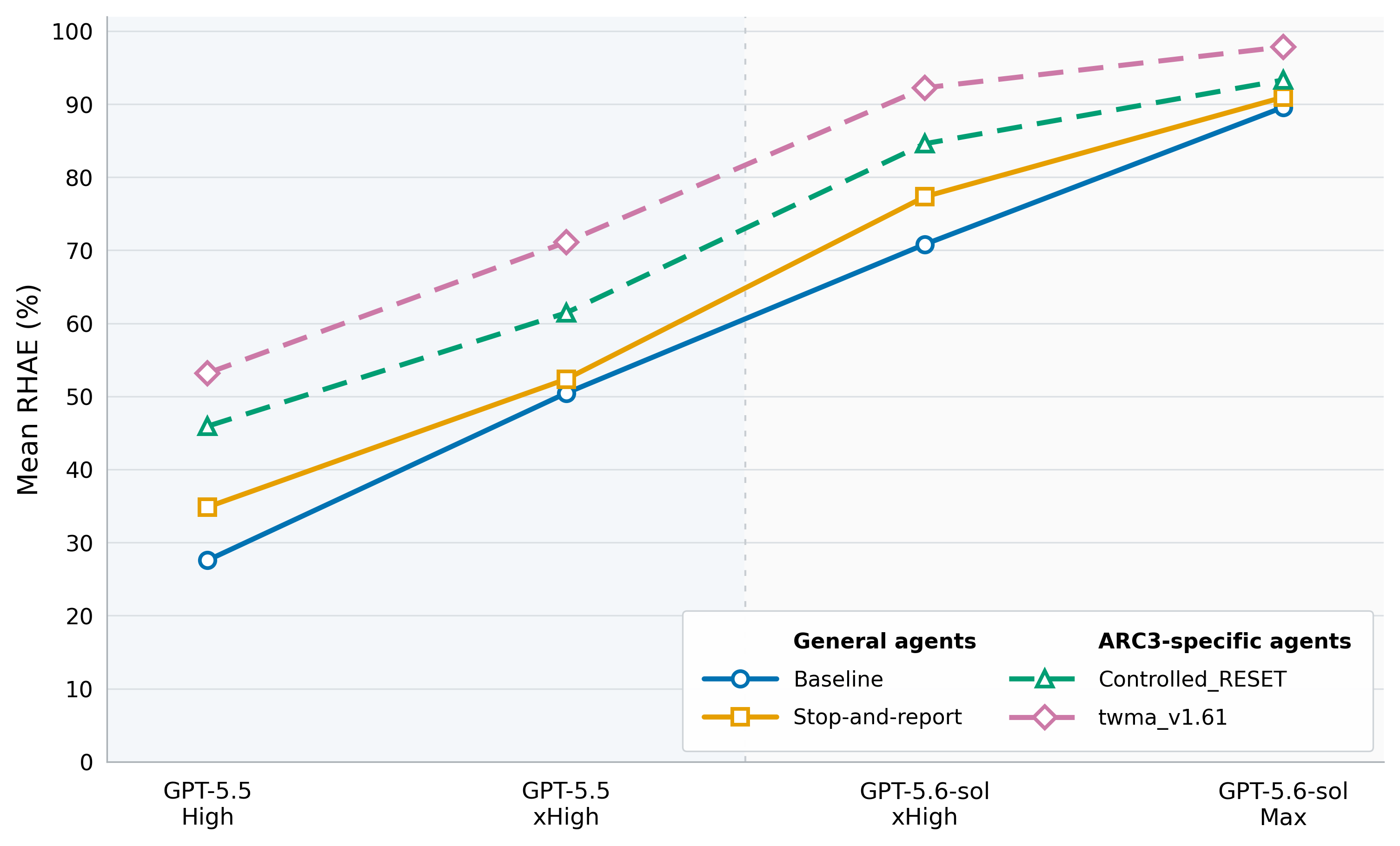}
\caption{Mean ARC-AGI-3 RHAE over retained evaluation blocks. Solid lines show the environment-agnostic
variants \AgentBaseline{} and \AgentStopReport{}; dashed lines show
\AgentControlledReset{} and \AgentTwma{}. The four horizontal positions are discrete
model--effort configurations, not a calibrated capability scale; connecting lines only
aid comparison. Each point averages three blocks, except \AgentTwma{} with GPT-5.6-sol
max, which averages two. Individual runs and exclusions are reported in
Appendix~\ref{app:arc3-individual}.}
\label{fig:arc3-results}
\end{figure}

Table~\ref{tab:arc3-results} and Figure~\ref{fig:arc3-results} compare the two
environment-agnostic variants with \AgentControlledReset{} and the \AgentTwma{}
reference defined in Section~\ref{sec:diagnostic-variants}. All results are from the
present campaign.

\begin{table}[!htbp]
\centering
\begin{minipage}{\linewidth}
\caption{ARC-AGI-3. Means over retained evaluation blocks of 25 independently played
games.}
\label{tab:arc3-results}
\centering
{\fontsize{10}{11.6}\selectfont
\setlength{\tabcolsep}{3.0pt}
\renewcommand{\arraystretch}{1}
\sbox{\compactmeasure}{%
\begin{tabular}{@{}ll@{\hspace{9pt}}S[table-format=2.2]S[table-format=2.2]@{\hspace{9pt}}S[table-format=2.2]S[table-format=2.2]@{\hspace{9pt}}S[table-format=2.2]S[table-format=2.2]@{\hspace{9pt}}S[table-format=2.2]S[table-format=2.2]@{}}
\toprule
 &  & \multicolumn{2}{c}{\AgentBaseline{}} & \multicolumn{2}{c}{\AgentStopReport{}} & \multicolumn{2}{c}{\AgentControlledReset{}} & \multicolumn{2}{c}{\AgentTwma{}} \\
\cmidrule(lr){3-4}\cmidrule(lr){5-6}\cmidrule(lr){7-8}\cmidrule(l){9-10}
Model & Effort & {RHAE $\uparrow$} & {Cost} & {RHAE $\uparrow$} & {Cost} & {RHAE $\uparrow$} & {Cost} & {RHAE $\uparrow$} & {Cost} \\
\midrule
GPT-5.5 & high & 27.57 & 9.36 & 34.86 & 14.19 & 45.92 & 42.19 & 53.19 & 64.09 \\
 & xhigh & 50.45 & 16.08 & 52.36 & 20.18 & 61.44 & 45.52 & 71.11 & 63.64 \\
\addlinespace[3pt]
GPT-5.6-sol & xhigh & 70.81 & 16.30 & 77.33 & 16.57 & 84.59 & 18.36 & 92.24 & 27.03 \\
 & max & 89.59 & 15.34 & 90.96 & 18.01 & 93.31 & 18.86 & 97.85 & 26.32 \\
\bottomrule
\end{tabular}}
\ifdim\wd\compactmeasure>\linewidth\PackageError{compacttables}{Table arc3_aggregate
exceeds text width}{}\fi
\usebox{\compactmeasure}}
\par\vspace{3pt}{\fontsize{9}{10.5}\selectfont\raggedright RHAE is in percent; cost is
in millions of normalized cost tokens (M). \AgentBaseline{} and \AgentStopReport{} are
environment-agnostic. \AgentControlledReset{} adds scripted reset management to
\AgentStopReport{}; it and \AgentTwma{} are ARC-AGI-3-specific. Means use three blocks,
except \AgentTwma{} with GPT-5.6-sol max, which uses two after the exclusions in
Appendix~\ref{app:web-search-exclusions}.\par}
\end{minipage}
\end{table}

\Needspace{6\baselineskip}
\paragraph{An environment-independent improvement.}
\AgentStopReport{} has higher mean RHAE than \AgentBaseline{} in all four model--effort
configurations, with gains of approximately 1.4--7.3 percentage points. Thus, changing
only the return-of-control instruction improves the mean score with both underlying
models, although the size of the gain varies across configurations.

\paragraph{A remaining specialization advantage.}
Mean scores increase from \AgentBaseline{} to \AgentStopReport{},
\AgentControlledReset{}, and \AgentTwma{} in every configuration. All four agents attain
their highest mean scores with GPT-5.6-sol max, but \AgentTwma{} still exceeds
\AgentStopReport{} by approximately seven RHAE points. The advantage of specialized
scaffolding therefore persists in the strongest evaluated configuration.

Adding scripted reset management recovers roughly one-third to three-fifths of the gap
between \AgentStopReport{} and \AgentTwma{}, without closing it. The reset convention is
already stated in the environment description (Appendix~\ref{app:arc-description}); the
intervention supplies control around available information rather than a new game rule.
The remaining differences from \AgentTwma{} are not separately ablated.

\paragraph{Performance and resource use.}
Within every model--effort configuration, the ordering of mean token-based cost matches
the ordering of mean RHAE across the four agents. \AgentStopReport{} costs approximately
2--52\% more than \AgentBaseline{}, and both specialized agents cost more still. The
performance gains are therefore accompanied by greater expenditure, even when the
implementation change is small.

\FloatBarrier
\subsection{AGI Maze: prompt modification and tutorial transfer}
\label{sec:results-maze}

\begin{figure}[tbp]
\centering
\includegraphics[width=\linewidth]{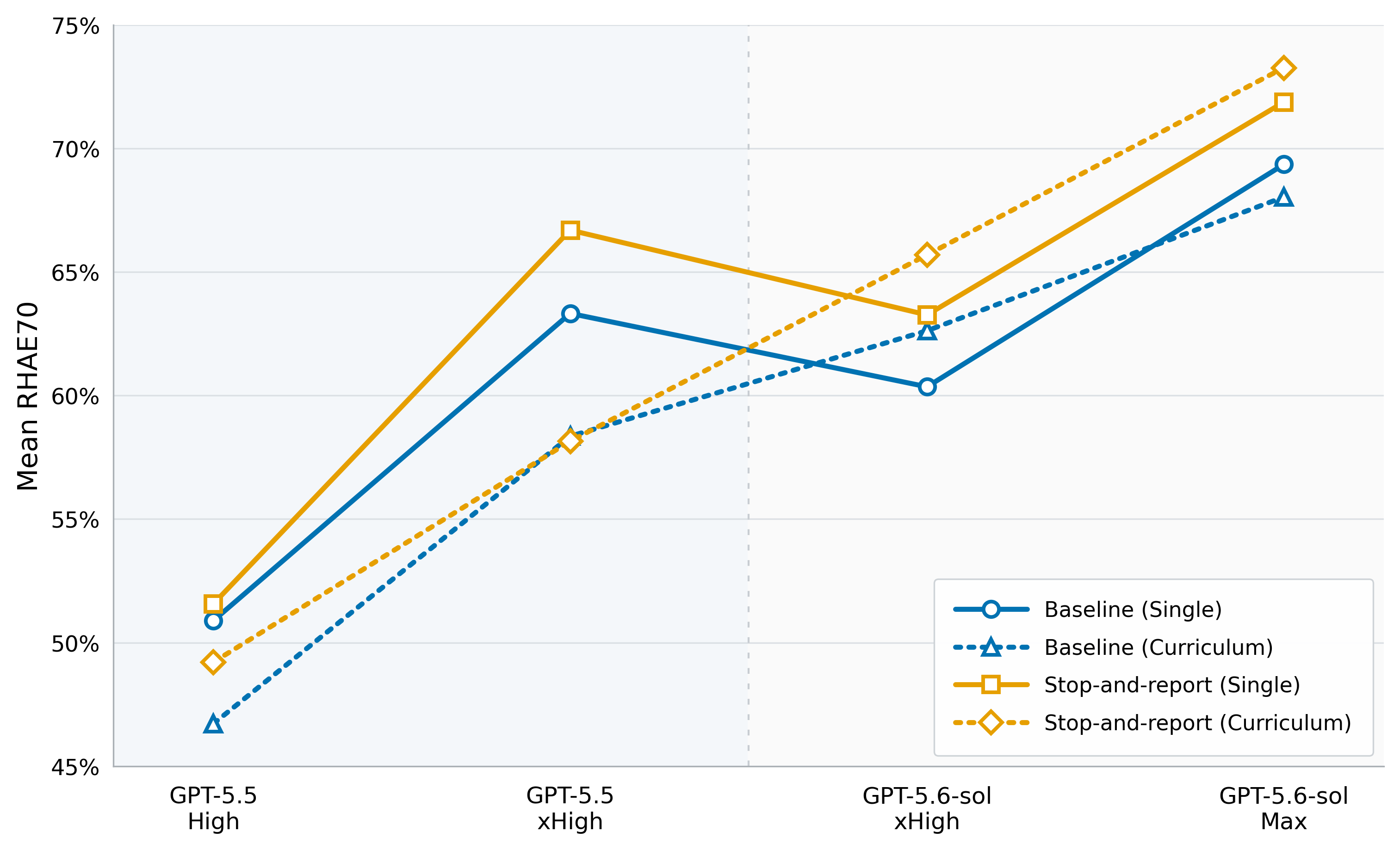}
\caption{AGI Maze mean RHAE70; higher is better. Solid lines show single-labyrinth
results, and dotted lines show the tutorial-set curriculum. Blue circles and triangles
denote \AgentBaseline{}; orange squares and diamonds denote \AgentStopReport{}. Each
point averages three complete 106-target evaluation blocks, except \AgentStopReport{}
with curriculum at GPT-5.5 xhigh and GPT-5.6-sol max, which average two. Model--effort
configurations are discrete. Individual execution batches and exclusions are given in
Appendix~\ref{app:maze-individual}.}
\label{fig:maze-results}
\end{figure}

Table~\ref{tab:maze-results} and Figure~\ref{fig:maze-results} separate two comparisons:
\AgentBaseline{} versus \AgentStopReport{} within each setting, and single-labyrinth
versus curriculum performance for each agent. RHAE70 is the score; the table also
retains x1-solved completion counts and token-based costs.

\begin{table}[!htbp]
\centering
\begin{minipage}{\linewidth}
\caption{AGI Maze. Means over retained evaluation blocks of 106 target labyrinths.}
\label{tab:maze-results}
\centering
{\fontsize{10}{11.6}\selectfont
\setlength{\tabcolsep}{3.4pt}
\renewcommand{\arraystretch}{1}
\sbox{\compactmeasure}{%
\begin{tabular}{@{}lll@{\hspace{10pt}}S[table-format=2.2]S[table-format=2.2]S[table-format=2.2]@{\hspace{12pt}}S[table-format=2.2]S[table-format=2.2]S[table-format=2.2]@{}}
\toprule
 &  &  & \multicolumn{3}{c}{\AgentBaseline{}} & \multicolumn{3}{c}{\AgentStopReport{}} \\
\cmidrule(lr){4-6}\cmidrule(l){7-9}
Model & Effort & Setting & {RHAE70 $\uparrow$} & {x1 solved} & {Cost} & {RHAE70 $\uparrow$} & {x1 solved} & {Cost} \\
\midrule
GPT-5.5 & high & Single & 50.89 & 54.00 & 14.08 & 51.57 & 50.00 & 17.31 \\
 &  & Curriculum & 46.72 & 48.33 & 21.67 & 49.21 & 49.00 & 27.95 \\
\addlinespace[3pt]
GPT-5.5 & xhigh & Single & 63.32 & 67.33 & 18.84 & 66.69 & 72.67 & 22.73 \\
 &  & Curriculum & 58.35 & 65.00 & 29.25 & 58.15 & 59.50 & 37.27 \\
\addlinespace[3pt]
GPT-5.6-sol & xhigh & Single & 60.35 & 62.67 & 8.55 & 63.26 & 69.00 & 9.14 \\
 &  & Curriculum & 62.62 & 67.67 & 21.84 & 65.69 & 74.67 & 30.16 \\
\addlinespace[3pt]
GPT-5.6-sol & max & Single & 69.35 & 74.67 & 8.28 & 71.87 & 81.00 & 10.78 \\
 &  & Curriculum & 68.03 & 74.67 & 27.95 & 73.26 & 85.00 & 39.00 \\
\bottomrule
\end{tabular}}
\ifdim\wd\compactmeasure>\linewidth\PackageError{compacttables}{Table maze_aggregate
exceeds text width}{}\fi
\usebox{\compactmeasure}}
\par\vspace{3pt}{\fontsize{9}{10.5}\selectfont\raggedright RHAE70 is reported as a
percentage; x1 solved is a descriptive count of targets completed within their original
reference budgets. Costs (M) and counts cover \texttt{0005}+\texttt{0009}. Curriculum
cost includes seven tutorials per target; only the final target is scored. Means use
three blocks, except \AgentStopReport{} with curriculum at GPT-5.5 xhigh and
GPT-5.6-sol max, which use two after the exclusions in
Appendix~\ref{app:web-search-exclusions}.\par}
\end{minipage}
\end{table}

\paragraph{Improvement beyond ARC-AGI-3.}
\AgentStopReport{} has higher mean RHAE70 in seven of eight comparisons, with gains
of up to 5.23 percentage points. The exception is GPT-5.5 xhigh with curriculum,
where \AgentStopReport{} shows a small decrease of 0.20 percentage points relative
to \AgentBaseline{}. The instruction was developed through ARC-AGI-3 experimentation,
without using AGI Maze performance to select it (Section~\ref{sec:evaluation-design}).
Its benefit therefore extends beyond the environment that motivated the change. Mean
token-based cost also increases in every comparison, by approximately 7--40\%.

\paragraph{Tutorial experience is not uniformly beneficial.}
Preceding tutorials reduce mean RHAE70 for both GPT-5.5-based agents at both effort
settings. With GPT-5.6-sol xhigh, tutorials improve both agents. At max, the effects are
smaller and mixed: tutorials lower \AgentBaseline{}'s score but raise
\AgentStopReport{}'s. The same curriculum can therefore help or hurt depending on the
model--effort configuration.

The curriculum also increases total token-based cost in every configuration. Additional
interaction and expenditure thus do not consistently translate into better target
performance. At xhigh effort, changing from GPT-5.5 to GPT-5.6-sol lowers
single-labyrinth RHAE70 for both agents but raises curriculum RHAE70, further
illustrating that the effect of a model change depends on the experience setting.

\FloatBarrier
\subsection{Terminal-Bench: performance beyond game environments}
\label{sec:results-terminal}

The same two agents also complete practical terminal tasks without a
Terminal-Bench-specific agent implementation. Table~\ref{tab:tb-results} reports mean
verifier reward; the GPT-5.6-sol configurations obtain approximately 84--87\%,
establishing task completion beyond the game settings.

\begin{table}[!htbp]
\centering
\begin{minipage}{.85\linewidth}
\caption{Terminal-Bench 2.1. Means over retained complete 89-task evaluation blocks.}
\label{tab:tb-results}
\centering
{\fontsize{10}{11.6}\selectfont
\setlength{\tabcolsep}{5pt}
\renewcommand{\arraystretch}{1}
\sbox{\compactmeasure}{%
\begin{tabular}{@{}ll@{\hspace{12pt}}S[table-format=2.2]S[table-format=2.2]@{\hspace{17pt}}S[table-format=2.2]S[table-format=2.2]@{}}
\toprule
 &  & \multicolumn{2}{c}{\AgentBaseline{}} & \multicolumn{2}{c}{\AgentStopReport{}} \\
\cmidrule(lr){3-4}\cmidrule(l){5-6}
Model & Effort & {Reward $\uparrow$} & {Cost} & {Reward $\uparrow$} & {Cost} \\
\midrule
GPT-5.5 & high & 79.40 & 5.80 & 80.34 & 5.85 \\
 & xhigh & 82.77 & 7.43 & 83.71 & 6.29 \\
\addlinespace[3pt]
GPT-5.6-sol & xhigh & 84.27 & 5.56 & 86.52 & 5.62 \\
 & max & 86.89 & 6.37 & 85.39 & 6.59 \\
\bottomrule
\end{tabular}}
\ifdim\wd\compactmeasure>\linewidth\PackageError{compacttables}{Table terminal_aggregate
exceeds text width}{}\fi
\usebox{\compactmeasure}}
\par\vspace{3pt}{\fontsize{9}{10.5}\selectfont\raggedright Reward is in percent; cost is
in M. Means use three blocks, except \AgentStopReport{} with GPT-5.5 at high and xhigh,
which use two after the exclusions in Appendix~\ref{app:web-search-exclusions}.\par}
\end{minipage}
\end{table}

On Terminal-Bench, \AgentStopReport{} produces small changes of mixed direction.
Differences from \AgentBaseline{} range from $-1.50$ to $+2.25$
percentage points across the four configurations. Mean token-based cost is lower for
\AgentStopReport{} only with GPT-5.5 xhigh and slightly higher in the other
configurations. Agent timeouts and cybersecurity-related interruptions are documented in
Appendices~\ref{app:tb-agent-timeouts} and~\ref{app:security-errors}.

\FloatBarrier
\Needspace{6\baselineskip}
\paragraph{Scope of the comparisons.}
\label{sec:results-scope}
The results concern the tested model--effort configurations and three environment
families. They compare performance under the reported execution policies, not at matched
inference cost. The small, mixed differences in mean Terminal-Bench scores do not
establish that \AgentStopReport{} is at least as effective as \AgentBaseline{}.
\Needspace{10\baselineskip}

\section{Discussion}
\label{sec:discussion}

The first research question concerns improving an environment-agnostic agent across
environments. \AgentStopReport{} improves mean ARC-AGI-3 RHAE and AGI Maze RHAE70, with
small, mixed changes on Terminal-Bench. These results show that a small change can
improve a general agent beyond the environment that motivated it and illustrate how to
test its effects elsewhere. The second question concerns why agents using the same
underlying LLM, with access to the same task information and general-purpose tools, can
nevertheless perform differently.

One hypothesis is that a sufficiently capable LLM could construct useful procedures for
itself, leaving little advantage to developer-supplied task-specific scaffolding. This
expectation is related to Sutton's \emph{The Bitter Lesson}, which emphasizes the
long-term value of general methods that exploit increasing
computation~\cite{sutton2019bitter}. Exgentic finds that backbone choice explains more
aggregate performance variation than architecture in its closed-source comparison,
although architecture remains consequential within a model~\cite{bandel2026general}. Our
previous ARC-AGI-3 study likewise found that gains from model capability and reasoning
effort often exceeded differences between the tested agent
variants~\cite{rodionov2026components}. These observations motivate the hypothesis; they
do not establish that scaffolding will become irrelevant.

The present results show that supplied procedures still matter with GPT-5.6-sol. Even at
max reasoning effort, \AgentStopReport{}, \AgentControlledReset{}, and \AgentTwma{}
achieve different mean ARC-AGI-3 scores. The specialization advantage therefore remains
in the evaluated setting. We consider two possible, complementary explanations: how
effectively the agent uses available information and procedures, and how much inference
it performs and where it allocates that inference.

\subsection{Using available information and procedures}
\label{sec:discussion-effective-use}

One possible explanation is that scaffolding helps the agent reconsider relevant
information or follow a useful procedure more reliably. When \AgentStopReport{} returns,
the controller reminds Codex of the goal and suggests rereading the environment
description (Appendix~\ref{app:continuation-prompt}). This may help it reconsider an
unsuccessful approach or revisit neglected information. The same reminder is available
to \AgentBaseline{}; the modification encourages Codex to return when it encounters
difficulty. It creates an opportunity for reconsideration rather than enforcing a
reflection procedure.

\AgentTwma{} illustrates the distinction between available information and prescribed
organization. The general agent receives information about levels and attempts and is
explicitly permitted to reorganize observations or create workspace files
(Appendices~\ref{app:arc-description} and~\ref{app:main-prompt}). It could therefore
build level-and-attempt indexes or maintain a textual account of the game. \AgentTwma{}
supplies the organization and requires that account. Its advantage need not come from
additional game facts; it may come from prescribing useful procedures instead of leaving
the agent to devise them. The comparison does not isolate which procedure contributes
most.

\AgentControlledReset{} provides a more focused example. The environment description
already tells the agent to reset after \texttt{GAME\_OVER} and avoid unnecessary resets
otherwise. Nevertheless, asking Codex to leave this decision to a scripted controller
improves mean performance. The intervention adds no new game rule; it changes how an
available rule is implemented. This does not mean that the scripted policy is optimal:
as discussed in Section~\ref{sec:controlled-reset}, a capable agent could sometimes save
actions by resetting an unrecoverable attempt before \texttt{GAME\_OVER}.

The curriculum experiment provides a complementary example: additional relevant
experience is not automatically beneficial. The tutorials give the agent practice
applying the maze mechanics, yet they reduce target performance for both GPT-5.5-based
variants at both effort settings. With GPT-5.6-sol, they help both variants at xhigh and
have smaller, mixed effects at max. The benefit of additional experience therefore
depends on the evaluated configuration, rather than following from its availability
alone.

\subsection{The amount and allocation of inference}
\label{sec:discussion-model-scaffold}
\label{sec:discussion-inference}

A second explanation is that the scaffold changes how much inference the agent performs
and when it performs it. Asking Codex to report a difficulty may prompt additional
reasoning while it formulates the report. The subsequent continuation lets it return to
the task. Together, these steps may increase inference at points where the agent is
uncertain, even with the model and reasoning-effort setting unchanged.

\AgentStopReport{} uses more normalized cost tokens than \AgentBaseline{} in every
ARC-AGI-3 and AGI Maze comparison. Within each ARC-AGI-3 configuration, the ordering
from \AgentBaseline{} through \AgentStopReport{} and \AgentControlledReset{} to
\AgentTwma{} is the same for mean score and mean cost. This pattern makes additional
inference a plausible contributor to the gains, but does not establish that greater
expenditure causes the improvement.

Cost tokens include weighted input and output, not only useful reasoning. Better
performance can also enable more expenditure: an agent reaching additional ARC-AGI-3
levels can continue working where another run has stopped. Performance and resource use
should therefore be considered together when interpreting the effect of
scaffolding~\cite{kapoor2024agents}. Better use of information can itself involve
additional inference, for example when the agent rereads a description or reasons about
how to reorganize its observations.

\subsection{Implications and future directions}
\label{sec:discussion-scope}
\label{sec:future-directions}

It remains unclear whether further gains should be sought primarily through better use
of available information and tools, through additional inference (for example, in a
multi-agent system), or through a combination of the two. Further analysis is needed to
determine their relative contributions and whether they can be meaningfully separated.

One direction is to improve the environment-agnostic agent through more explicit
procedures, including adaptations of the verification and simplification ideas explored
in our earlier work~\cite{rodionov2026components}. The objective is to obtain further
gains on some environments without material regressions on others, while retaining one
unchanged agent implementation.

A complementary direction is to expand the general capabilities initially available to
the agent. Examples include reusable task-general skills, support for symbolic reasoning
through a system such as MeTTa~\cite{hyperon2026metta}, or theorem-proving tools. These
additions would supply tools or procedural knowledge, rather than only change the
guidance for using existing tools, but could still be provided uniformly across
environments.
\Needspace{25\baselineskip}

\section{Conclusions}
\label{sec:conclusions}

We evaluated one Codex-based agent in two nearly identical prompt variants, each
deployed unchanged across ARC-AGI-3, AGI Maze, and Terminal-Bench. \AgentStopReport{}
improves mean scores in all four ARC-AGI-3 configurations and seven of eight AGI Maze
comparisons, while Terminal-Bench shows small, mixed changes. The comparison demonstrates that a small,
environment-independent modification can improve performance beyond the environment that
motivated it, although the performance gains come with higher token-based cost.

Stronger models may eventually construct useful procedures for themselves and reduce the
advantage of developer-supplied scaffolding. Our results show that this advantage
remains with GPT-5.6-sol. The \AgentTwma{} reference prescribes organization and
procedures that the general agent has the information and tools to construct, and
\AgentControlledReset{} recovers part of its advantage by implementing a convention
already stated in the environment description. The tutorial experiment adds a
complementary observation: access to additional relevant experience does not uniformly
improve performance.

The unresolved question is whether scaffolding helps mainly through more effective use
of information and tools, additional inference, or both. The present experiments do not
separate these effects, which may reinforce one another. EnvTraverse allows new agent
variants to be evaluated across environments and compared with \AgentBaseline{}. This
supports further research on improving environment-agnostic agents through better
procedures or additional general capabilities, while assessing performance gains,
possible regressions, and changes in inference cost.
\Needspace{6\baselineskip}

\section*{Acknowledgements}

The author thanks Alexey Potapov and Vita Potapova for their assistance with AGI Maze,
for clarifying the reference action budgets, and for sharing the internal playthrough
results that informed the RHAE70 reference factor.
\clearpage
\bibliographystyle{unsrtnat}
\bibliography{references}
\clearpage
\appendix

\section{Environment-Provided Information}
\label{app:environment-inputs}

This appendix reproduces the environment descriptions, templates, and messages from the
source code used in the experiments. The source code is available through the
\href{https://github.com/astroseger/envtraverse}{EnvTraverse project page}. Quoted text
is reproduced verbatim, with line wrapping adjusted for typesetting. Template fields and
returned observation files are explained where needed.

\subsection{ARC-AGI-3 description}
\label{app:arc-description}

This description is supplied at initialization for every ARC-AGI-3 game. Its ``Provided
Artifacts'' section specifies the observation files, and its ``Provided Helper
Programs'' section describes \texttt{ascii\_to\_png.py}, supplied in the initial
workspace.

\sourceinput{prompts/environments/arc3/description.md}

\subsection{AGI Maze single-labyrinth description}
\label{app:maze-single}

\paragraph{Template.}
The environment description is generated from the following template, shown with simple
placeholder names. All information substituted into these fields is supplied by AGI
Maze, including the public map.

\sourceinput{prompts/environments/agimaze_single/description_template.txt}

\paragraph{Rendered example.}
The following is a complete environment description for one non-tutorial labyrinth.

\sourceinput{prompts/environments/agimaze_single/description_example.txt}

\subsection{AGI Maze set description and transition messages}
\label{app:maze-set}

\subsubsection{Environment description}
\label{app:maze-set-description}

The environment description is:

\sourceinput{prompts/environments/agimaze_set/description.md}

The initial observation is \texttt{message.md} containing:

\sourceinput{prompts/environments/agimaze_set/initial_message.txt}

The environment waits for \texttt{\{"id":"start"\}}; it does not begin the first
labyrinth automatically.

\subsubsection{When a labyrinth starts}
\label{app:maze-set-start}

A valid \texttt{start} action returns \texttt{labyrinth\_description.md}, containing the
environment description used for that labyrinth in the single-labyrinth setting. It also
returns \texttt{state.txt}, containing the same state information as in the
single-labyrinth environment, with the labyrinth index added. Tutorial labyrinths
additionally provide AGI Maze's current public map and player position in
\texttt{map.txt}. Subsequent actions update the state and, for tutorials, the map;
non-tutorial labyrinths omit \texttt{map.txt}.

\subsubsection{Between labyrinths}
\label{app:maze-set-transitions}

\paragraph{After a labyrinth is solved and another remains.}
The response replaces the ordinary observation files with \texttt{message.md}
containing:

\sourceinput{prompts/environments/agimaze_set/next_after_success.txt}

\paragraph{After the action limit is reached and another labyrinth remains.}
When the current labyrinth is not solved within its budget, \texttt{message.md} instead
contains:

\sourceinput{prompts/environments/agimaze_set/next_after_failure.txt}

In both cases the environment remains nonterminal and waits for another \texttt{start}
action.

\subsubsection{At the end of the set}
\label{app:maze-set-final}

\paragraph{After the final labyrinth is solved.}
The terminal response contains \texttt{message.md} with:

\sourceinput{prompts/environments/agimaze_set/final_success.txt}

\paragraph{After the final labyrinth reaches its action limit.}
The terminal response instead contains:

\sourceinput{prompts/environments/agimaze_set/final_failure.txt}

\subsection{Terminal-Bench description template}
\label{app:terminal-description}

The following template is supplied at initialization after replacing \texttt{\{budget\}}
with the resolved wall-clock solving budget in seconds and \texttt{\{instruction\}} with
the task's public instruction. Its file-retrieval limit is 33,554,432 bytes in the
supplied configuration. The template explains the six available action categories,
terminal timing, and the separation of the task filesystem from the agent's workspace.

\sourceinput{prompts/environments/terminal_bench/description_template.md}

\subsubsection{Returned observation files}
\label{app:terminal-observations}

The following filenames describe the actual responses; they are not additional
instructions appended to the template. Reading a terminal returns \texttt{screen.txt} or
\texttt{history.txt}, together with \texttt{terminal.json} describing the terminal and
capture. Creating a terminal returns its initial screen and metadata. The initial
environment observation additionally includes \texttt{session.json}. Sending terminal
input returns \texttt{input.json}, acknowledging input delivery rather than command
completion.

File retrieval returns the original bytes in \texttt{file.bin} and metadata in
\texttt{file.json}. Upload returns \texttt{transfer.json}, acknowledging the source,
destination, and copied contents. The workspace-access convention is described in
Section~\ref{sec:runtime}; transfer implementation details are provided in the project
documentation. When solving ends, the response contains a terminal flag and
\texttt{message.txt}, not verifier scores or test diagnostics. Verification is still run
after the solving time limit expires.

\section{Agent Prompts and Their Composition}
\label{app:agent-prompts}

\subsection{Shared main and continuation templates}
\label{app:shared-prompts}

\subsubsection{Main prompt}
\label{app:main-prompt}

The main prompt below is identical in \AgentBaseline{} and \AgentStopReport{}. It gives
the location of the environment-owned description, rather than embedding the description
of a particular environment into the agent prompt.

\sourceinput{prompts/agents/main_prompt.md}

\subsubsection{Common continuation text}
\label{app:continuation-prompt}

The common continuation prefix is:
\sourceinput{prompts/agents/continuation.txt}

\subsubsection{No-action reminder}
\label{app:stuck-reminder}

The additional no-action reminder is:
\sourceinput{prompts/agents/no_action_reminder.txt}

\subsection{Variant strings and prompt composition}
\label{app:variant-strings}

\subsubsection{\AgentBaseline{} instruction}
\label{app:baseline-instruction}

For \AgentBaseline{}, the variant string is:
\sourceinput{prompts/agents/baseline_instruction.txt}

\subsubsection{\AgentStopReport{} instruction}
\label{app:stop-instruction}

For \AgentStopReport{}, the variant string is:
\sourceinput{prompts/agents/stop_and_report_instruction.txt}

\subsubsection{Prompt composition}
\label{app:prompt-composition}

The controller composes the initial input from the main prompt, the selected variant
string, and the text printed by the client when it starts the session. The latter lists
the environment-description and observation locations. A normal continuation consists of
the common continuation prefix followed by the variant string. The no-action branch
appends the shared reminder to that same continuation. Newline separators are preserved
in the source templates. Thus, the intervention is present at initialization and on
later continuations; it is not applied only after the first return.

\clearpage

\section{Individual Runs and Reporting Details}
\label{app:individual-results}

This appendix reports all completed individual runs, including those excluded from the
aggregate analysis. The implementation identifiers are \texttt{agent\_v1} for \AgentBaseline{},
\texttt{agent\_v1\_01\_wstop} for \AgentStopReport{}, and
\texttt{test\_arc3\_agent\_v1\_02\_wstop\_wreset} for \AgentControlledReset{};
\AgentTwma{} uses its implementation identifier as its paper name. The first two
variants are environment-agnostic. Run indices identify repetitions within a
configuration; a matching index across different agents does not establish a shared
random seed.

The complete results and run artifacts supporting this study are available through the
\href{https://github.com/astroseger/envtraverse}{EnvTraverse project page}.

\subsection{Identification and exclusion of affected evaluation blocks}
\label{app:web-search-exclusions}

We searched the preserved execution logs for the substring
\texttt{search\_service.web\_run} to identify attempted calls to the web search tool
described in Section~\ref{sec:apps-search-exclusions}, including unsuccessful calls.
Every evaluation block containing a task execution flagged by this audit was excluded
in full from the aggregate analysis, irrespective of the call's success or the task's
outcome. The audit identified 15 task executions: three
in AGI Maze, one in ARC-AGI-3, and eleven in Terminal-Bench. These occurred in five
evaluation blocks, listed below using the configuration names and run numbers in
Tables~\ref{tab:arc3-individual}--\ref{tab:tb-individual}, together with the task
identifiers that triggered exclusion.

\begin{enumerate}
\item \textbf{ARC-AGI-3, \AgentTwma{}, GPT-5.6-sol / max, Run 3.}
The flagged game was \texttt{bp35}.

\item \textbf{AGI Maze, \AgentStopReport{}, GPT-5.5 / xhigh, Curriculum, Run 2.}
The flagged task was \path{CLASSIC_MEDIUM_rivers_7x7} in batch
\texttt{0009}. Both batches, \texttt{0005} and \texttt{0009}, were excluded,
although batch \texttt{0005} contained no flagged execution.

\item \textbf{AGI Maze, \AgentStopReport{}, GPT-5.6-sol / max, Curriculum, Run 3.}
The flagged tasks were \path{EXTENDED_boat_STAGE-03-5x5} in batch
\texttt{0005} and \path{CLASSIC_EASY_rivers_4x4} in batch
\texttt{0009}. Both batches were excluded.

\item \textbf{Terminal-Bench, \AgentStopReport{}, GPT-5.5 / high, Run 1.}
The flagged tasks were
\path{break-filter-js-from-html},
\path{build-pov-ray},
\path{dna-assembly},
\path{dna-insert},
\path{mteb-leaderboard}, and
\path{protein-assembly}.

\item \textbf{Terminal-Bench, \AgentStopReport{}, GPT-5.5 / xhigh, Run 1.}
The flagged tasks were
\path{build-pov-ray},
\path{compile-compcert},
\path{mcmc-sampling-stan},
\path{mteb-leaderboard}, and
\path{protein-assembly}.
\end{enumerate}

A Maze evaluation block comprises both \texttt{0005} and \texttt{0009} batches,
so both batches are excluded whenever either contains a flagged execution.
All scores and costs from these blocks are omitted
from the aggregate analysis; no task-level replacement or imputation is used. For
example, the web search attempt during \texttt{bp35} in \AgentTwma{} with
GPT-5.6-sol max led to exclusion of the entire evaluation block labeled Run 3.
The reported mean score and cost therefore use only the two retained blocks,
labeled Runs 1 and 2.

The individual tables below retain the original observed values for excluded blocks
and mark them with asterisks.

\subsection{ARC-AGI-3 individual runs}
\label{app:arc3-individual}
\begin{table}[H]
\centering
\begin{minipage}{\linewidth}
\caption{ARC-AGI-3 individual runs. All 48 completed runs, each containing 25 games,
including the excluded block.}
\label{tab:arc3-individual}
\centering
{\fontsize{10}{11.6}\selectfont
\setlength{\tabcolsep}{3.8pt}
\renewcommand{\arraystretch}{1}
\sbox{\compactmeasure}{%
\begin{tabular}{@{}l@{\hspace{10pt}}S[table-format=2.2]S[table-format=2.0]S[table-format=2.2]@{\hspace{10pt}}S[table-format=2.2]S[table-format=2.0]S[table-format=2.2]@{\hspace{10pt}}S[table-format=2.2,table-space-text-post={\textsuperscript{*}}]S[table-format=2.0,table-space-text-post={\textsuperscript{*}}]S[table-format=2.2,table-space-text-post={\textsuperscript{*}}]@{}}
\toprule
 & \multicolumn{3}{c}{Run 1} & \multicolumn{3}{c}{Run 2} & \multicolumn{3}{c}{Run 3} \\
\cmidrule(lr){2-4}\cmidrule(lr){5-7}\cmidrule(l){8-10}
Agent & {RHAE} & {Unsolved} & {Cost} & {RHAE} & {Unsolved} & {Cost} & {RHAE} & {Unsolved} & {Cost} \\
\midrule
\multicolumn{10}{@{}l}{\textbf{GPT-5.5 / high}} \\
\AgentBaseline{} & 29.54 & 21 & 9.01 & 30.26 & 19 & 10.45 & 22.91 & 22 & 8.62 \\
\AgentStopReport{} & 33.33 & 17 & 12.26 & 37.06 & 17 & 16.89 & 34.19 & 18 & 13.41 \\
\AgentControlledReset{} & 48.57 & 14 & 41.87 & 44.96 & 14 & 34.95 & 44.23 & 14 & 49.73 \\
\AgentTwma{} & 52.44 & 13 & 66.42 & 53.98 & 12 & 63.12 & 53.15 & 15 & 62.72 \\
\addlinespace[3pt]
\multicolumn{10}{@{}l}{\textbf{GPT-5.5 / xhigh}} \\
\AgentBaseline{} & 50.22 & 15 & 15.71 & 52.52 & 15 & 15.39 & 48.60 & 16 & 17.13 \\
\AgentStopReport{} & 58.48 & 11 & 27.00 & 49.48 & 15 & 17.03 & 49.13 & 15 & 16.52 \\
\AgentControlledReset{} & 61.77 & 9 & 52.26 & 58.03 & 10 & 43.73 & 64.51 & 8 & 40.55 \\
\AgentTwma{} & 70.09 & 7 & 56.73 & 69.74 & 7 & 62.47 & 73.52 & 7 & 71.73 \\
\addlinespace[3pt]
\multicolumn{10}{@{}l}{\textbf{GPT-5.6-sol / xhigh}} \\
\AgentBaseline{} & 68.98 & 7 & 18.24 & 76.92 & 5 & 12.55 & 66.53 & 6 & 18.10 \\
\AgentStopReport{} & 75.88 & 5 & 14.88 & 81.93 & 4 & 14.67 & 74.19 & 4 & 20.17 \\
\AgentControlledReset{} & 87.51 & 1 & 18.86 & 85.65 & 2 & 18.76 & 80.61 & 4 & 17.46 \\
\AgentTwma{} & 93.88 & 1 & 24.89 & 90.94 & 1 & 30.44 & 91.91 & 0 & 25.76 \\
\addlinespace[3pt]
\multicolumn{10}{@{}l}{\textbf{GPT-5.6-sol / max}} \\
\AgentBaseline{} & 87.68 & 3 & 16.55 & 92.21 & 2 & 14.35 & 88.87 & 3 & 15.12 \\
\AgentStopReport{} & 92.35 & 2 & 17.22 & 87.29 & 2 & 18.80 & 93.23 & 0 & 18.00 \\
\AgentControlledReset{} & 92.13 & 2 & 17.81 & 94.81 & 0 & 17.96 & 92.99 & 0 & 20.81 \\
\AgentTwma{} & 97.51 & 0 & 24.20 & 98.19 & 1 & 28.44 & 97.88{\textsuperscript{*}} & 0{\textsuperscript{*}} & 27.18{\textsuperscript{*}} \\
\bottomrule
\end{tabular}}
\ifdim\wd\compactmeasure>\linewidth\PackageError{compacttables}{Table arc3_individual
exceeds text width}{}\fi
\usebox{\compactmeasure}}
\par\vspace{3pt}{\fontsize{9}{10.5}\selectfont\raggedright RHAE is in percent; cost is
in M. Unsolved counts games not fully completed; RHAE still credits partial progress.
Run numbers identify independent repetitions, not matched seeds across agents.
\textsuperscript{*}Original observed values from an evaluation block excluded from
the aggregate analysis; see Appendix~\ref{app:web-search-exclusions}.\par}
\end{minipage}
\end{table}

\clearpage
\subsection{AGI Maze individual execution batches}
\label{app:maze-individual}

The table reports the \texttt{0005} and \texttt{0009} execution batches separately, each
containing 53 target labyrinths. The main analysis combines the two batches at each
repetition into a 106-target evaluation block. RHAE70 and x1-solved counts use the
definitions in Section~\ref{sec:evaluation-measures}. Only the declared \texttt{CLASSIC}
and \texttt{EXTENDED} targets contribute to the reported results and costs. All 96
completed batches are shown; 92 batches form the 46 retained evaluation blocks.

\begin{table}[H]
\centering
\begin{minipage}{\linewidth}
\caption{AGI Maze individual execution batches. All 96 completed batches, each
containing 53 targets, including both batches of each excluded block.}
\label{tab:maze-individual}
\centering
{\fontsize{9.5}{11.1}\selectfont
\setlength{\tabcolsep}{2.35pt}
\renewcommand{\arraystretch}{1}
\sbox{\compactmeasure}{%
\begin{tabular}{@{}l@{\hspace{7pt}}l@{\hspace{9pt}}S[table-format=2.2]S[table-format=2.0]S[table-format=1.2]S[table-format=2.2]@{\hspace{9pt}}S[table-format=2.2,table-space-text-post={\textsuperscript{*}}]S[table-format=2.0,table-space-text-post={\textsuperscript{*}}]S[table-format=1.2,table-space-text-post={\textsuperscript{*}}]S[table-format=2.2,table-space-text-post={\textsuperscript{*}}]@{\hspace{9pt}}S[table-format=2.2,table-space-text-post={\textsuperscript{*}}]S[table-format=2.0,table-space-text-post={\textsuperscript{*}}]S[table-format=1.2,table-space-text-post={\textsuperscript{*}}]S[table-format=2.2,table-space-text-post={\textsuperscript{*}}]@{}}
\toprule
 &  & \multicolumn{4}{c}{Run 1} & \multicolumn{4}{c}{Run 2} & \multicolumn{4}{c}{Run 3} \\
\cmidrule(lr){3-6}\cmidrule(lr){7-10}\cmidrule(l){11-14}
Agent & Batch & {RHAE70} & {x1} & {Frac.} & {Cost} & {RHAE70} & {x1} & {Frac.} & {Cost} & {RHAE70} & {x1} & {Frac.} & {Cost} \\
\midrule
\multicolumn{14}{@{}l}{\textbf{GPT-5.5 / high --- Single}} \\
\AgentBaseline{} & \texttt{0005} & 57.34 & 32 & 1.49 & 6.71 & 54.31 & 26 & 1.58 & 5.39 & 53.67 & 29 & 1.47 & 7.60 \\
 & \texttt{0009} & 45.07 & 23 & 1.71 & 6.54 & 48.50 & 28 & 1.58 & 7.52 & 46.46 & 24 & 1.79 & 8.49 \\
\AgentStopReport{} & \texttt{0005} & 52.93 & 27 & 1.46 & 8.05 & 51.91 & 28 & 1.52 & 9.75 & 58.24 & 32 & 1.38 & 7.56 \\
 & \texttt{0009} & 48.02 & 22 & 1.53 & 7.69 & 44.77 & 18 & 1.56 & 12.08 & 53.51 & 23 & 1.13 & 6.80 \\
\addlinespace[3pt]
\multicolumn{14}{@{}l}{\textbf{GPT-5.5 / high --- Curriculum}} \\
\AgentBaseline{} & \texttt{0005} & 49.57 & 25 & 1.82 & 11.66 & 50.76 & 28 & 1.89 & 11.55 & 51.67 & 26 & 1.77 & 10.64 \\
 & \texttt{0009} & 40.85 & 19 & 2.00 & 10.35 & 43.70 & 22 & 1.88 & 10.69 & 43.75 & 25 & 1.84 & 10.12 \\
\AgentStopReport{} & \texttt{0005} & 50.70 & 26 & 1.70 & 13.57 & 52.57 & 26 & 1.57 & 14.27 & 51.23 & 29 & 1.63 & 15.51 \\
 & \texttt{0009} & 45.21 & 22 & 1.73 & 12.12 & 44.91 & 20 & 1.72 & 15.00 & 50.63 & 24 & 1.53 & 13.37 \\
\addlinespace[3pt]
\multicolumn{14}{@{}l}{\textbf{GPT-5.5 / xhigh --- Single}} \\
\AgentBaseline{} & \texttt{0005} & 67.67 & 37 & 1.19 & 9.63 & 63.60 & 35 & 1.25 & 10.63 & 66.51 & 34 & 1.14 & 7.79 \\
 & \texttt{0009} & 61.37 & 33 & 1.11 & 8.83 & 60.58 & 29 & 1.20 & 10.50 & 60.21 & 34 & 1.10 & 9.13 \\
\AgentStopReport{} & \texttt{0005} & 63.66 & 38 & 1.21 & 15.08 & 68.51 & 37 & 1.04 & 13.98 & 72.84 & 39 & 0.89 & 9.93 \\
 & \texttt{0009} & 63.50 & 36 & 1.02 & 9.53 & 66.15 & 34 & 1.08 & 8.91 & 65.50 & 34 & 1.00 & 10.76 \\
\addlinespace[3pt]
\multicolumn{14}{@{}l}{\textbf{GPT-5.5 / xhigh --- Curriculum}} \\
\AgentBaseline{} & \texttt{0005} & 60.99 & 34 & 1.33 & 13.73 & 62.37 & 35 & 1.28 & 15.15 & 62.58 & 34 & 1.30 & 14.61 \\
 & \texttt{0009} & 54.69 & 31 & 1.51 & 14.47 & 54.17 & 31 & 1.50 & 15.24 & 55.30 & 30 & 1.51 & 14.56 \\
\AgentStopReport{} & \texttt{0005} & 57.77 & 31 & 1.34 & 19.72 & 65.49{\textsuperscript{*}} & 37{\textsuperscript{*}} & 1.31{\textsuperscript{*}} & 18.00{\textsuperscript{*}} & 60.22 & 31 & 1.57 & 20.08 \\
 & \texttt{0009} & 59.34 & 30 & 1.26 & 19.09 & 54.24{\textsuperscript{*}} & 25{\textsuperscript{*}} & 1.26{\textsuperscript{*}} & 17.13{\textsuperscript{*}} & 55.26 & 27 & 1.15 & 15.66 \\
\addlinespace[3pt]
\multicolumn{14}{@{}l}{\textbf{GPT-5.6-sol / xhigh --- Single}} \\
\AgentBaseline{} & \texttt{0005} & 63.96 & 34 & 1.18 & 4.21 & 62.73 & 34 & 1.28 & 4.48 & 60.98 & 33 & 1.30 & 4.24 \\
 & \texttt{0009} & 60.15 & 30 & 1.13 & 3.57 & 53.08 & 25 & 1.36 & 4.85 & 61.20 & 32 & 1.10 & 4.29 \\
\AgentStopReport{} & \texttt{0005} & 68.90 & 39 & 1.05 & 4.68 & 67.12 & 37 & 1.09 & 4.55 & 62.76 & 36 & 1.07 & 4.26 \\
 & \texttt{0009} & 59.93 & 31 & 1.08 & 4.72 & 60.47 & 29 & 1.11 & 4.56 & 60.37 & 35 & 1.01 & 4.64 \\
\addlinespace[3pt]
\multicolumn{14}{@{}l}{\textbf{GPT-5.6-sol / xhigh --- Curriculum}} \\
\AgentBaseline{} & \texttt{0005} & 68.34 & 39 & 1.02 & 9.68 & 67.81 & 38 & 1.18 & 11.19 & 66.12 & 33 & 0.99 & 10.21 \\
 & \texttt{0009} & 61.81 & 32 & 1.02 & 11.02 & 59.96 & 30 & 1.04 & 10.28 & 51.68 & 31 & 1.27 & 13.13 \\
\AgentStopReport{} & \texttt{0005} & 66.01 & 38 & 1.16 & 16.02 & 66.79 & 36 & 0.94 & 14.94 & 69.46 & 41 & 0.98 & 14.29 \\
 & \texttt{0009} & 66.68 & 40 & 0.95 & 14.84 & 66.46 & 37 & 0.98 & 13.85 & 58.74 & 32 & 1.22 & 16.53 \\
\addlinespace[3pt]
\multicolumn{14}{@{}l}{\textbf{GPT-5.6-sol / max --- Single}} \\
\AgentBaseline{} & \texttt{0005} & 69.45 & 37 & 0.99 & 4.08 & 71.60 & 40 & 0.85 & 3.76 & 74.36 & 39 & 0.88 & 3.95 \\
 & \texttt{0009} & 68.24 & 39 & 0.90 & 3.63 & 64.78 & 32 & 0.97 & 4.57 & 67.69 & 37 & 1.03 & 4.86 \\
\AgentStopReport{} & \texttt{0005} & 74.90 & 43 & 0.82 & 4.74 & 75.52 & 44 & 0.79 & 5.08 & 73.87 & 42 & 0.83 & 4.67 \\
 & \texttt{0009} & 73.68 & 40 & 0.82 & 4.93 & 68.88 & 40 & 0.96 & 6.47 & 64.37 & 34 & 0.95 & 6.45 \\
\addlinespace[3pt]
\multicolumn{14}{@{}l}{\textbf{GPT-5.6-sol / max --- Curriculum}} \\
\AgentBaseline{} & \texttt{0005} & 71.17 & 41 & 0.95 & 13.97 & 68.87 & 34 & 0.89 & 13.93 & 70.08 & 41 & 0.89 & 13.31 \\
 & \texttt{0009} & 69.73 & 37 & 0.88 & 14.24 & 67.56 & 38 & 0.92 & 13.47 & 60.75 & 33 & 1.00 & 14.93 \\
\AgentStopReport{} & \texttt{0005} & 76.64 & 45 & 0.81 & 19.12 & 73.00 & 43 & 0.83 & 16.89 & 73.27{\textsuperscript{*}} & 40{\textsuperscript{*}} & 0.83{\textsuperscript{*}} & 19.82{\textsuperscript{*}} \\
 & \texttt{0009} & 72.10 & 42 & 0.87 & 22.39 & 71.28 & 40 & 0.84 & 19.60 & 71.57{\textsuperscript{*}} & 37{\textsuperscript{*}} & 0.88{\textsuperscript{*}} & 18.34{\textsuperscript{*}} \\
\bottomrule
\end{tabular}}
\ifdim\wd\compactmeasure>\linewidth\PackageError{compacttables}{Table maze_individual
exceeds text width}{}\fi
\usebox{\compactmeasure}}
\par\vspace{3pt}{\fontsize{9}{10.5}\selectfont\raggedright RHAE70 is in percent;
\textit{x1} is the number solved within the original reference budget. \textit{Frac.} is
the mean of \texttt{steps/max\_steps} for solved targets, with 5 assigned to unsolved
targets. Cost (M) is per 53-target batch and includes tutorials in curriculum runs; only
the final targets are scored. Batch identifiers are retained separately. Run indices do
not imply matched seeds across agents. \textsuperscript{*}Original observed values from
an evaluation block excluded from the aggregate analysis; both batches are marked even
when only one contained a flagged task. See Appendix~\ref{app:web-search-exclusions}.\par}
\end{minipage}
\end{table}

\clearpage
\subsection{Terminal-Bench individual runs}
\label{app:tb-individual}
\begin{table}[H]
\centering
\begin{minipage}{\linewidth}
\caption{Terminal-Bench individual runs. All 24 completed runs, each containing 89
tasks, including the two excluded blocks.}
\label{tab:tb-individual}
\centering
{\fontsize{10}{11.6}\selectfont
\setlength{\tabcolsep}{3.8pt}
\renewcommand{\arraystretch}{1}
\sbox{\compactmeasure}{%
\begin{tabular}{@{}l@{\hspace{10pt}}S[table-format=2.2,table-space-text-post={\textsuperscript{*}}]S[table-format=2.0,table-space-text-post={\textsuperscript{*}}]S[table-format=2.2,table-space-text-post={\textsuperscript{*}}]@{\hspace{10pt}}S[table-format=2.2]S[table-format=2.0]S[table-format=2.2]@{\hspace{10pt}}S[table-format=2.2]S[table-format=2.0]S[table-format=2.2]@{}}
\toprule
 & \multicolumn{3}{c}{Run 1} & \multicolumn{3}{c}{Run 2} & \multicolumn{3}{c}{Run 3} \\
\cmidrule(lr){2-4}\cmidrule(lr){5-7}\cmidrule(l){8-10}
Agent & {Reward} & {Unsolved} & {Cost} & {Reward} & {Unsolved} & {Cost} & {Reward} & {Unsolved} & {Cost} \\
\midrule
\multicolumn{10}{@{}l}{\textbf{GPT-5.5 / high}} \\
\AgentBaseline{} & 77.53 & 20 & 6.61 & 79.78 & 18 & 5.15 & 80.90 & 17 & 5.65 \\
\AgentStopReport{} & 82.02{\textsuperscript{*}} & 16{\textsuperscript{*}} & 6.11{\textsuperscript{*}} & 79.78 & 18 & 5.60 & 80.90 & 17 & 6.10 \\
\addlinespace[3pt]
\multicolumn{10}{@{}l}{\textbf{GPT-5.5 / xhigh}} \\
\AgentBaseline{} & 85.39 & 13 & 7.41 & 82.02 & 16 & 7.35 & 80.90 & 17 & 7.52 \\
\AgentStopReport{} & 79.78{\textsuperscript{*}} & 18{\textsuperscript{*}} & 8.09{\textsuperscript{*}} & 79.78 & 18 & 6.70 & 87.64 & 11 & 5.89 \\
\addlinespace[3pt]
\multicolumn{10}{@{}l}{\textbf{GPT-5.6-sol / xhigh}} \\
\AgentBaseline{} & 85.39 & 13 & 6.18 & 87.64 & 11 & 5.22 & 79.78 & 18 & 5.28 \\
\AgentStopReport{} & 85.39 & 13 & 6.42 & 86.52 & 12 & 4.73 & 87.64 & 11 & 5.70 \\
\addlinespace[3pt]
\multicolumn{10}{@{}l}{\textbf{GPT-5.6-sol / max}} \\
\AgentBaseline{} & 85.39 & 13 & 6.12 & 86.52 & 12 & 6.74 & 88.76 & 10 & 6.24 \\
\AgentStopReport{} & 83.15 & 15 & 6.82 & 87.64 & 11 & 6.52 & 85.39 & 13 & 6.42 \\
\bottomrule
\end{tabular}}
\ifdim\wd\compactmeasure>\linewidth\PackageError{compacttables}{Table
terminal_individual exceeds text width}{}\fi
\usebox{\compactmeasure}}
\par\vspace{3pt}{\fontsize{9}{10.5}\selectfont\raggedright Reward is in percent; cost is
in M. Unsolved counts zero-reward outcomes, including verifier timeouts. Run indices do
not imply matched seeds across agents. \textsuperscript{*}Original observed values from
an evaluation block excluded from the aggregate analysis; see
Appendix~\ref{app:web-search-exclusions}.\par}
\end{minipage}
\end{table}

\subsection{Terminal-Bench agent timeouts}
\label{app:tb-agent-timeouts}

Across the 22 retained Terminal-Bench evaluation blocks, 19 task executions reached the
solving time limit. Verification was run afterward, and all 19 received reward zero.

\subsection{Cybersecurity-related errors and task outcomes}
\label{app:security-errors}

In some GPT-5.6-sol task executions, Codex returned the following cybersecurity-related
error:
\begin{quote}
\small
This content was flagged for possible cybersecurity risk. If this seems wrong, try
rephrasing your request. To get authorized for security work, join the Trusted Access
for Cyber program: \url{https://chatgpt.com/cyber}
\end{quote}
The run summaries report 18 task executions encountering such an error and 12 ending
with a cybersecurity-related recovery failure. All 12 stopped task executions had used
the ten allowed recovery attempts. Table~\ref{tab:security-outcomes} reports the
affected evaluation blocks and the resulting verifier outcomes.

\begin{table}[H]
\centering
\begin{minipage}{\linewidth}
\caption{Cybersecurity-related errors in GPT-5.6-sol runs. Only runs with reported
errors are shown. Counts refer to task executions.}
\label{tab:security-outcomes}
\centering
{\fontsize{10}{11.6}\selectfont
\setlength{\tabcolsep}{5pt}
\renewcommand{\arraystretch}{1}
\sbox{\compactmeasure}{%
\begin{tabular}{@{}llr rr rr@{}}
\toprule
Agent & Effort & {Run} & {\shortstack{Error\\encountered}} & {\shortstack{Recovery\\failed}} & \multicolumn{2}{c}{Stopped tasks} \\
\cmidrule(l){6-7}
 & & & & & {Reward 0} & {Reward 1} \\
\midrule
\AgentBaseline{} & xhigh & 1 & 4 & 2 & 2 & 0 \\
\AgentBaseline{} & max & 1 & 3 & 3 & 3 & 0 \\
\AgentBaseline{} & max & 3 & 3 & 2 & 2 & 0 \\
\AgentStopReport{} & xhigh & 1 & 3 & 3 & 2 & 1 \\
\AgentStopReport{} & max & 1 & 5 & 2 & 2 & 0 \\
\midrule
Total & & & 18 & 12 & 11 & 1 \\
\bottomrule
\end{tabular}}
\ifdim\wd\compactmeasure>\linewidth\PackageError{compacttables}{Table security_outcomes
exceeds text width}{}\fi
\usebox{\compactmeasure}}
\par\vspace{3pt}{\fontsize{9}{10.5}\selectfont\raggedright The final two columns give
verifier outcomes for stopped tasks. Every stopped task in these runs had ten recorded
recovery attempts.\par}
\end{minipage}
\end{table}

\clearpage

\section{Motivation for the RHAE70 Reference Factor}
\label{app:rhae70-reference}

According to the AGI Maze team, the benchmark's \texttt{max\_steps} budgets were defined
with some slack. The team shared results from internal playthroughs covering eight
labyrinths, conducted by one practiced player who knew the mechanics but had not
previously seen those labyrinths. Four labyrinths had an action budget of 180 and four
had a budget of 230. The mean per-labyrinth ratio of actions used to budget was
approximately 0.617. Seven playthroughs used fewer than 70 percent of the budget; the
remaining ratio was $162/230\approx0.704$. These observations informed our choice of 0.7
as a rounded reference factor for awarding full credit.

These internal tests were not designed to estimate representative human performance or
calibrate a reference across all labyrinth types. RHAE70 is an author-defined,
RHAE-inspired score using a proxy action reference, not a measured human-baseline score.
A value of 100 means that the target was solved within the chosen threshold; it does not
establish human-level or superhuman performance. The score saturates at that threshold
and does not distinguish faster solutions.

\paragraph{Score selection and sensitivity.}
RHAE70 was adopted during analysis and manuscript preparation, after examining mean step
fractions, to give the reported benchmark scores a consistent higher-is-better
direction. The internal playthrough results shared by the AGI Maze team informed the
choice of reference factor. We also examined mean step fraction and analogous scores
with reference factors of 0.6 and 0.8. The broad comparison favors \AgentStopReport{},
but for GPT-5.5 xhigh with curriculum the ordering depends on normalization:
RHAE60 favors \AgentStopReport{}, at 47.94\% versus 47.74\%, whereas RHAE80 favors
\AgentBaseline{}, at 67.14\% versus 65.79\%. These comparisons use the retained blocks
after the exclusions in Appendix~\ref{app:web-search-exclusions}. RHAE70 is the score
used in the main comparisons.
\end{document}